\documentclass[lettersize,journal]{IEEEtran}
\usepackage{amsmath,amsfonts}
\usepackage{algorithmic}
\usepackage{algorithm}
\usepackage{array}
\usepackage[caption=false,font=normalsize,labelfont=sf,textfont=sf]{subfig}
\usepackage{textcomp}
\usepackage{utfsym}
\usepackage{stfloats}
\usepackage{url}
\usepackage{tabularx,booktabs}
\usepackage{tikz}
\usepackage{makecell} 
\usepackage{fontawesome}

\usetikzlibrary{shapes,arrows,positioning}
\newcommand{\tabincell}[1]{\begin{tabular}[c]{@{}l@{}}#1\end{tabular}}
\usepackage{verbatim}
\usepackage{graphicx}
\usepackage{cite}
\usepackage{booktabs}
\usepackage{multicol}
\usepackage{multirow}
\usepackage{tcolorbox}
\usepackage{colortbl}
\usepackage{util}
\usepackage{array}
\usepackage{longtable}
\usepackage{orcidlink}

\begin{document}

\title{Large Foundation Models for Trajectory Prediction in Autonomous Driving: A Comprehensive Survey}

\author{
    Wei Dai$^{\orcidlink{0009-0008-9877-6363}}$,
    Shengen Wu$^{\orcidlink{0009-0001-3432-3583}}$,
    Wei Wu$^{\orcidlink{0009-0005-5381-781X}}$,
    Zhenhao Wang, Sisuo Lyu, Haicheng Liao, \\
    Limin Yu, Weiping Ding,~\IEEEmembership{Senior Member,~IEEE}, Runwei Guan$^{\orcidlink{0000-0003-4013-2107}}$, Yutao Yue$^{\orcidlink{0000-0003-4532-0924}}$,~\IEEEmembership{Senior Member,~IEEE}
\thanks{This work was supported in part by the XJTLU Postgraduate Research Scholarship under Grant FOS2104JP07, in part by the Red Bird MPhil Program at the Hong Kong University of Science and Technology (Guangzhou). (Corresponding author: Yutao Yue.)}%
\thanks{Wei Dai is with the Department of Mathematical Sciences, School of Physical sciences, University of Liverpool, L69 3BX Liverpool, U.K. and with the Department of Communications and Networking, School of Advanced Technology, Xi’an Jiaotong-Liverpool University, Suzhou 215000, China. }%
\thanks{Shengen Wu, Wei Wu and Runwei Guan are with the Thrust of Artificial Intelligence, The Hong Kong University of Science and Technology (Guangzhou), Guangzhou 511400, China.}%
\thanks{Zhenhao Wang is with the Deep Interdisciplinary Intelligence Lab, The Hong Kong University of Science and Technology (Guangzhou) as a research intern, Guangzhou 511400, China, and with the School of Mathematics and Statistics, Shandong University, Weihai 264209, China.}
\thanks{Sisuo Lyu is with the Thrust of Data Science and Analytics, The Hong Kong University of Science and Technology (Guangzhou), Guangzhou 511400, China.}%
\thanks{Weiping Ding is with the School of Artificial Intelligence and Computer Science, Nantong University, Nantong 226019, China.}%
\thanks{Limin Yu is with the Department of Communications and Networking, School of Advanced Technology, Xi’an Jiaotong-Liverpool University, Suzhou 215000, China.}%
\thanks{Yutao Yue is with the Thrust of Artificial Intelligence, the Thrust of Intelligent Transportation and the Deep Interdisciplinary Intelligence Lab, The Hong Kong University of Science and Technology (Guangzhou), Guangzhou 511400, China, also with the Institute of Deep Perception Technology, Jiangsu Industrial Technology Research Institute, Wuxi 214028, China(e-mail: yutaoyue@hkust-gz.edu.cn).}
}

\markboth{September~2025}%
{Shell \MakeLowercase{\textit{et al.}}: A Sample Article Using IEEEtran.cls for IEEE Journals}


\maketitle

\begin{abstract}
Trajectory prediction serves as a critical functionality in autonomous driving, enabling the anticipation of future motion paths for traffic participants such as vehicles and pedestrians, which is essential for driving safety. Although conventional deep learning methods have improved accuracy, they remain hindered by inherent limitations, including lack of interpretability, heavy reliance on large-scale annotated data, and weak generalization in long-tail scenarios. The rise of Large Foundation Models (LFMs) is transforming the research paradigm of trajectory prediction. This survey offers a systematic review of recent advances in LFMs, particularly Large Language Models (LLMs) and Multimodal Large Language Models (MLLMs) for trajectory prediction. By integrating linguistic and scene semantics, LFMs facilitate interpretable contextual reasoning, significantly enhancing prediction safety and generalization in complex environments. The article highlights three core methodologies: trajectory-language mapping, multimodal fusion, and constraint-based reasoning. It covers prediction tasks for both vehicles and pedestrians, evaluation metrics, and dataset analyses. Key challenges such as computational latency, data scarcity, and real-world robustness are discussed, along with future research directions including low-latency inference, causality-aware modeling, and motion foundation models.
\end{abstract}


\section{Introduction}

\begin{figure}
    \centering
    \includegraphics[width=0.9\linewidth]{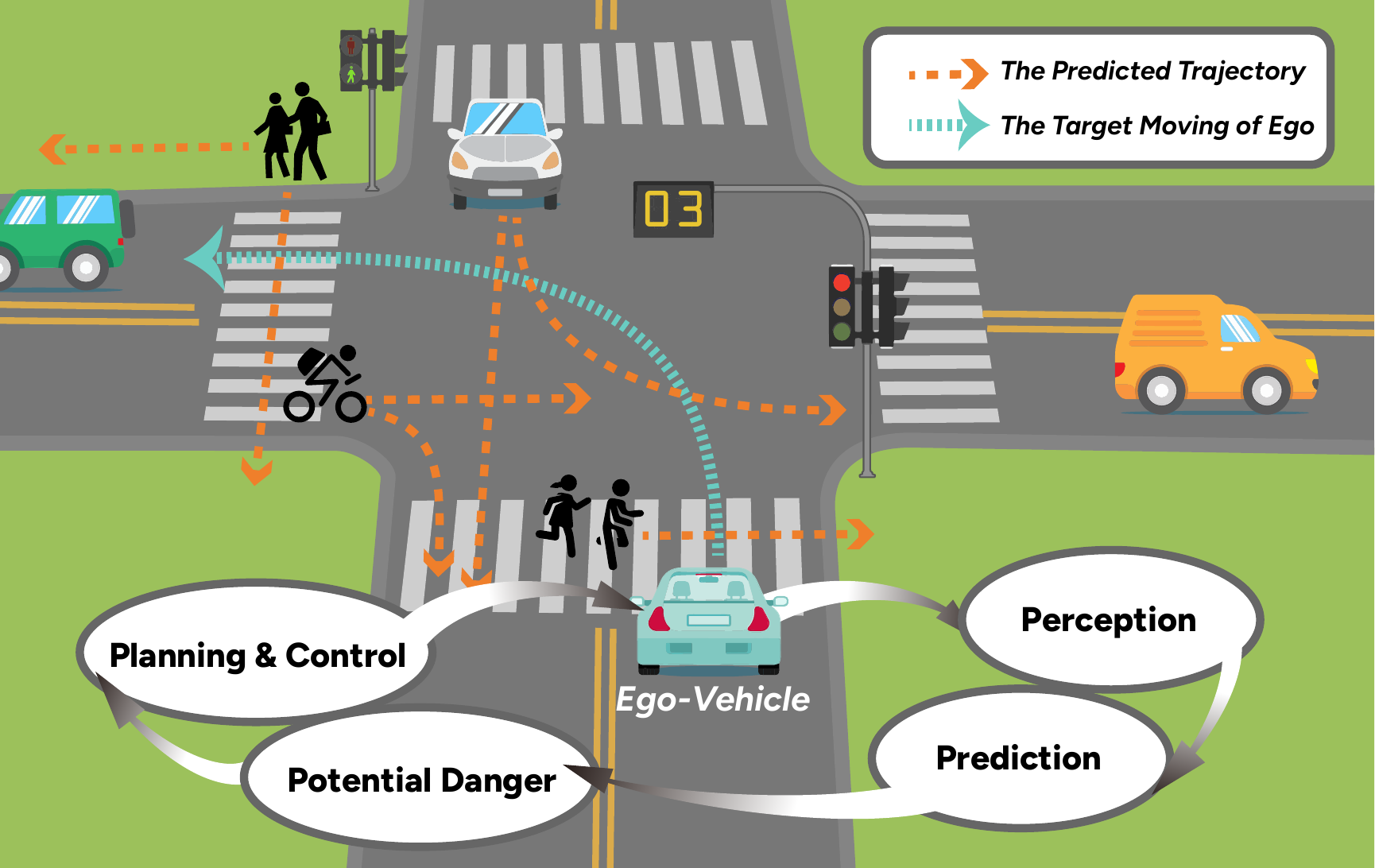}
    \vspace{-2mm}
    \caption{Interaction and Trajectory Prediction for Ego-Vehicle in Autonomous Driving. The figure illustrates how an Ego-Vehicle navigates a complex intersection using a "Perception-Prediction-Planning \& Control" loop. The orange dashed arrows represent the predicted trajectories of other traffic participants (e.g., pedestrians, cyclists, cars, and vans), while the green dotted arrow indicates the target movement of the Ego-Vehicle.}
    \label{fig:placeholder}
\end{figure}

\begin{figure}[t!]
    \centering
    \includegraphics[scale=0.55]{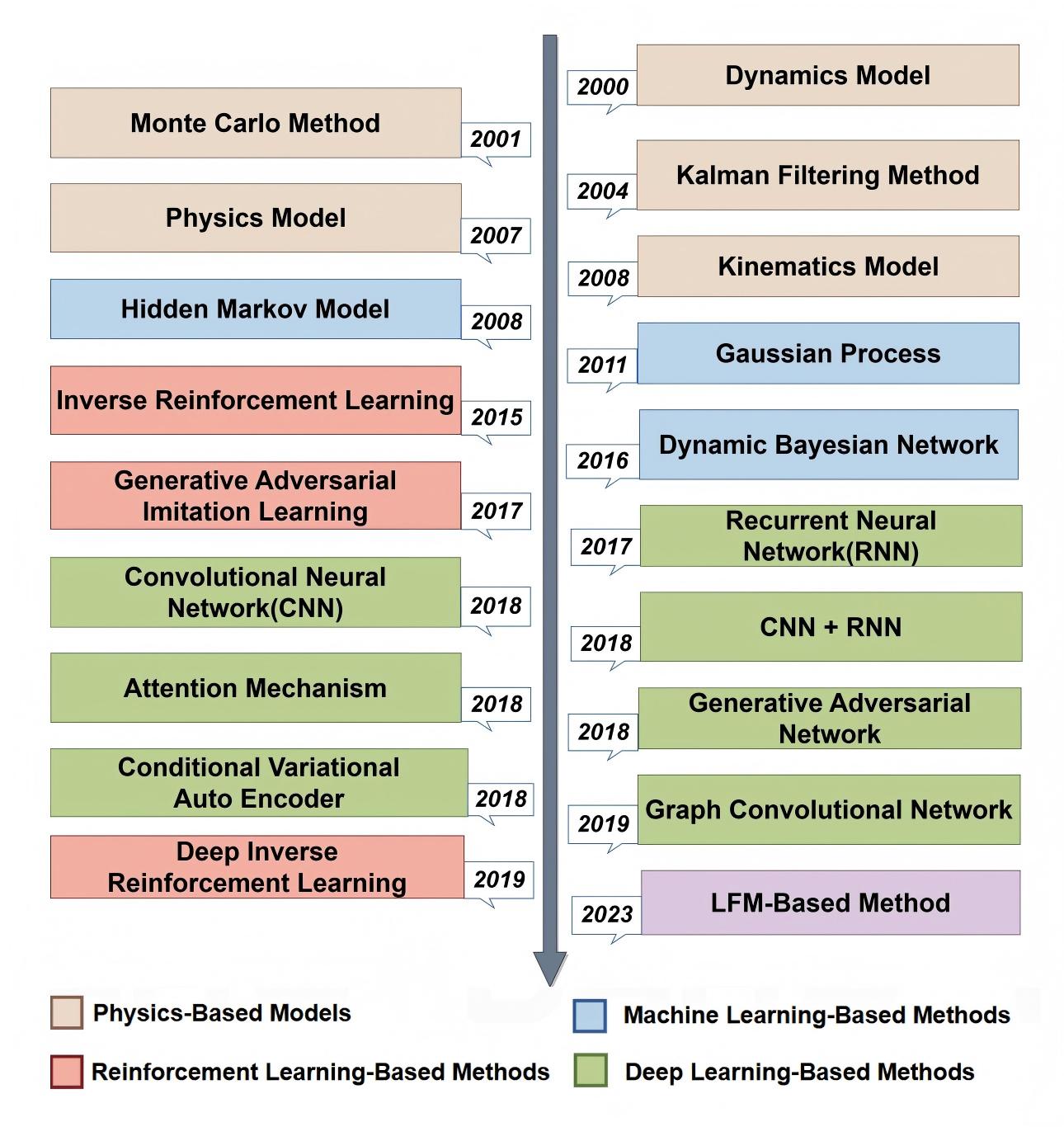}
    \vspace{-4mm}
    \caption{Timeline of Trajectory Prediction Methodologies} 
    \label{fig:timeline}
    \vspace{-4mm}
\end{figure}

\IEEEPARstart{T}{he} pursuit of fully autonomous driving hinges on a system's ability to anticipate the future actions of surrounding traffic participants, a task known as trajectory prediction. The field witnessed a paradigm shift with the advent of deep learning, which provided a powerful data-driven approach to this problem. Architectures based on Recurrent Neural Networks (RNNs), Graph Neural Networks (GNNs) and generative models became the dominant paradigm, demonstrating remarkable success in learning complex spatiotemporal patterns directly from data \cite{tang2019multiple, ding2019online, messaoud2021trajectory, kuo2022trajectory, li2019grip, feng2019vehicle}. These models excelled in capturing nuanced interactions and significantly extended the accuracy of long-term predictions. Despite these achievements, a critical challenge remained in their lack of an interpretable decision-making process, posing a significant barrier to formal verification and safety certification \cite{zablocki2022explainability}. Furthermore, their performance was highly dependent on large quantities of annotated data, and their ability to generalize to novel long-tailed scenarios not well represented in training datasets remained a significant concern \cite{li2022uqnet}. These limitations provided the impetus to explore new paradigms capable of more robust and human-like reasoning. To provide a clear overview of this paradigm shift in trajectory prediction, Fig.~\ref{fig:timeline} outlines the historical evolution of methodologies, from early rule-based systems to contemporary semantic reasoning approaches powered by Large Foundation Models (LFMs).

\begin{figure*}[b!]
    \centering
    \includegraphics[width=\linewidth]{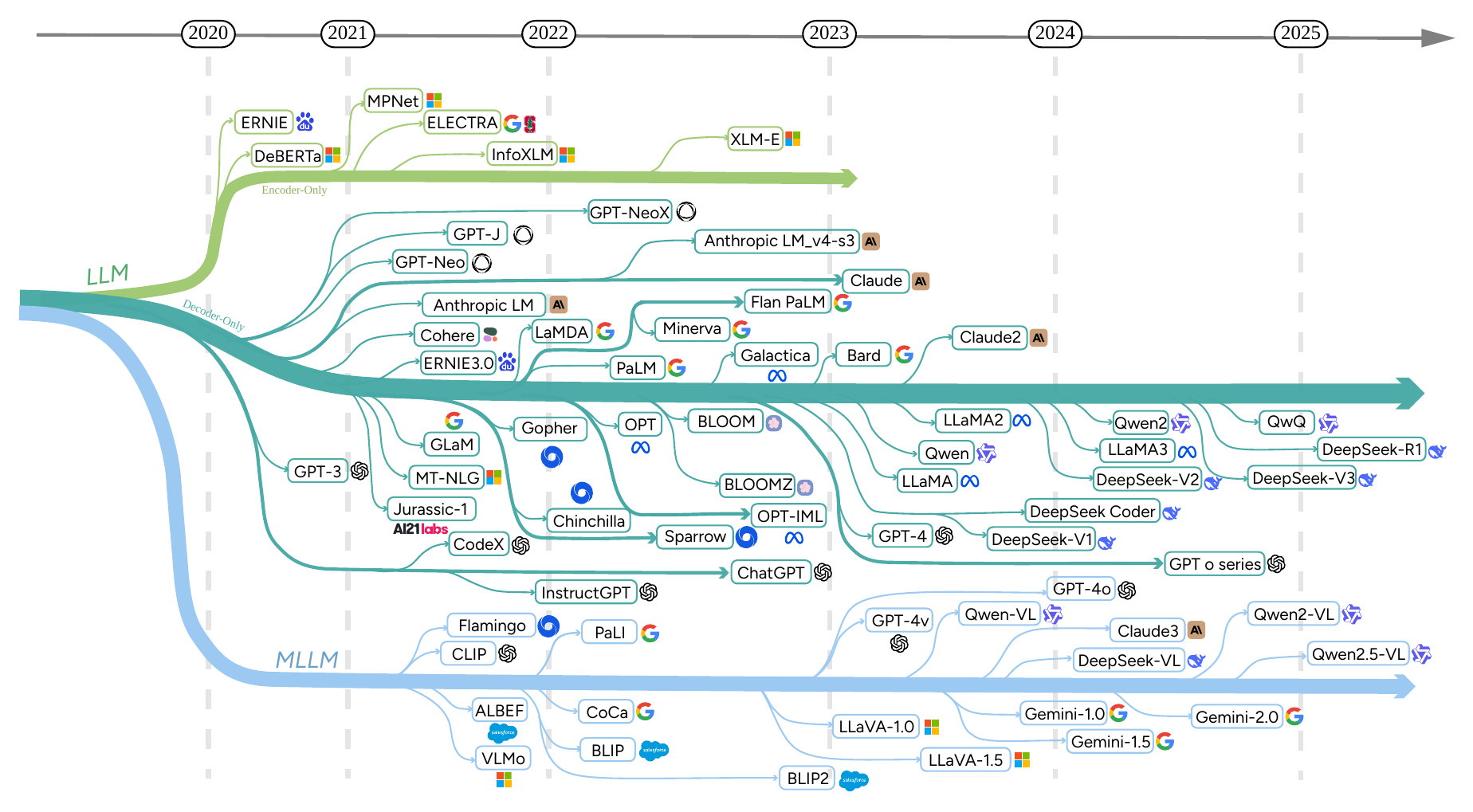}
    \vspace{-4mm}
    \caption{An Overview of the Development Road map of Large Foundation Models in Recent Years.} 
    \label{fig:llm-roadmap}
    \vspace{-4mm}
\end{figure*}

The advent of LFMs introduces a novel paradigm for addressing the fundamental limitations of traditional trajectory prediction frameworks. LFMs broadly refer to a spectrum of pre-trained models that serve as general purpose computational backbones across various modalities. This continuum spans from earlier modality-specific encoders, such as vision and text encoders trained on large-scale datasets, to more recent Large Language Models (LLMs) and Multimodal Large Language Models (MLLMs) that exhibit strong reasoning, generation, and alignment capabilities. As illustrated in Fig.~\ref{fig:llm-roadmap}, the path of LFM development reflects an ongoing shift toward increasingly integrated, knowledge-rich, and generalizable architectures. This shift fundamentally transforms trajectory prediction from a low-level pattern recognition task to one grounded in semantic understanding and cognitive reasoning. Using the vast repositories of world knowledge embedded within them, LLMs can internalize and apply common sense principles, traffic regulations, and social conventions to the prediction task \cite{yangLLM4DriveSurveyLarge2024, xuLargeReasoningModels2025}. MLLMs further extend these capabilities by synergistically integrating heterogeneous data streams from cameras and LiDAR with textual instructions, allowing a holistic understanding of the driving scene \cite{wangComprehensiveReviewMultimodal2024}. Crucially, through techniques such as Chain-of-Thought (CoT) reasoning, these models can articulate the step-by-step rationale behind their predictions in natural language, providing a new layer of transparency that addresses the challenge of uninterpretable decision making \cite{cuiChainofThoughtAutonomousDriving2025}. Recent surveys have also emphasized that the effectiveness of foundation models in autonomous driving depends on four core capabilities: generalized knowledge, spatial understanding, multisensor robustness, and temporal reasoning\cite{sathyam2025foundation}. Although these capabilities have been systematically explored in the context of perception tasks, their specific integration and evaluation in trajectory prediction remain underexplored. Given the surge in research applying these advanced models to trajectory prediction, a systematic survey is of great importance. Although the fields of autonomous driving trajectory prediction and LLMs have been extensively reviewed separately, a dedicated survey of their intersection is still lacking. On the one hand, reviews on trajectory prediction predominantly cover methods prior to the advent of LFMs. However, reviews on LLMs often explore their broader applications without a dedicated analysis of their use in trajectory prediction. This survey aims to bridge this gap by providing a comprehensive review that focuses specifically on the application of LFMs for trajectory prediction.

To this end, we review the current landscape of language foundation models in trajectory prediction, with a focus on recent technological trends. Our main contributions are:
\begin{itemize}
    \item We present a comprehensive survey on the application of LLMs and MLLMs to trajectory prediction, establishing a structured taxonomy that categorizes existing studies into core methodologies: trajectory-language mapping, multimodal fusion, and constraint-based reasoning.
    \item We consolidate the mainstream prediction tasks for both vehicles and pedestrians, along with their corresponding widely-used evaluation metrics.
    \item We summarize and analyze existing benchmark datasets relevant to language-enhanced trajectory prediction.
    \item We provide an in-depth discussion of the benefits, persistent challenges, and research gaps in this domain, while also offering insights into future trends such as model distillation for real-time deployment and causal inference for safety verification.
\end{itemize}

The remainder of this paper is organized as follows: Section II provides a comprehensive background on trajectory prediction, covering problem formulation, traditional methods, deep learning, machine learning, and reinforcement learning approaches for vehicle and pedestrian trajectory prediction. Section III delves into the core of LLM-based trajectory prediction, including perception and scene understanding, vehicle trajectory prediction with LLMs, pedestrian trajectory prediction with LLMs and multi-agent trajectory prediction with LLMs. Section IV discusses experimental benchmarks, evaluation metrics, and performance comparisons between traditional and LLM-based trajectory prediction methods. Section V presents a discussion of the advantages, challenges, and future directions of LLM-based trajectory prediction. Finally, the key conclusions are presented in Section VI.

\begin{figure}[t!]
    \centering
    \includegraphics[width=\linewidth]{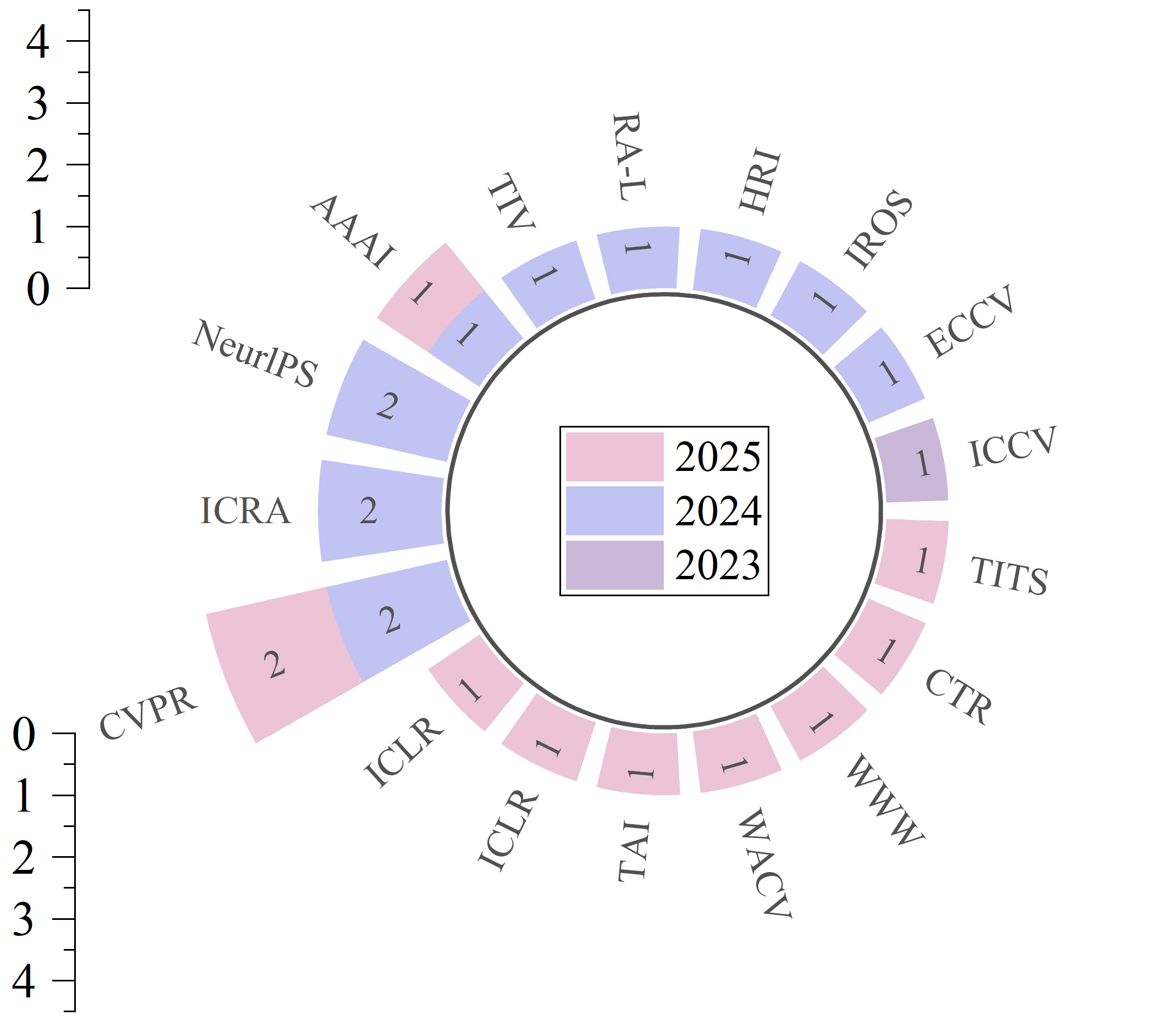}
    \vspace{-4mm}
    \caption{Distribution of Research Publications on LLM-based Trajectory Prediction in Top-Tier Conferences and Journals(2023-2025.09), all abbreviations follow standard usage (see Table\ref{tab:abbreviations})}
    \label{fig:publications}
    \vspace{-4mm}
\end{figure}

\section{Trajectory Prediction}
Trajectory prediction constitutes a core technology within autonomous driving systems, with the aim of inferring future motion paths of dynamic traffic participants (e.g.,
vehicles and pedestrians) based on historical observational data (such as position and velocity) and contextual knowledge (including maps and traffic rules) \cite{huang2022survey}. This field is broadly categorized into vehicle trajectory prediction and pedestrian trajectory prediction, differentiated by predicted entities. Methodologically, approaches are further classified into traditional methods and data-driven methods, the latter encompassing deep learning and reinforcement learning techniques \cite{ding2023incorporating, golchoubian2023pedestrian}.

\begin{figure*}[t!]
    \centering
    \includegraphics[width=\linewidth]{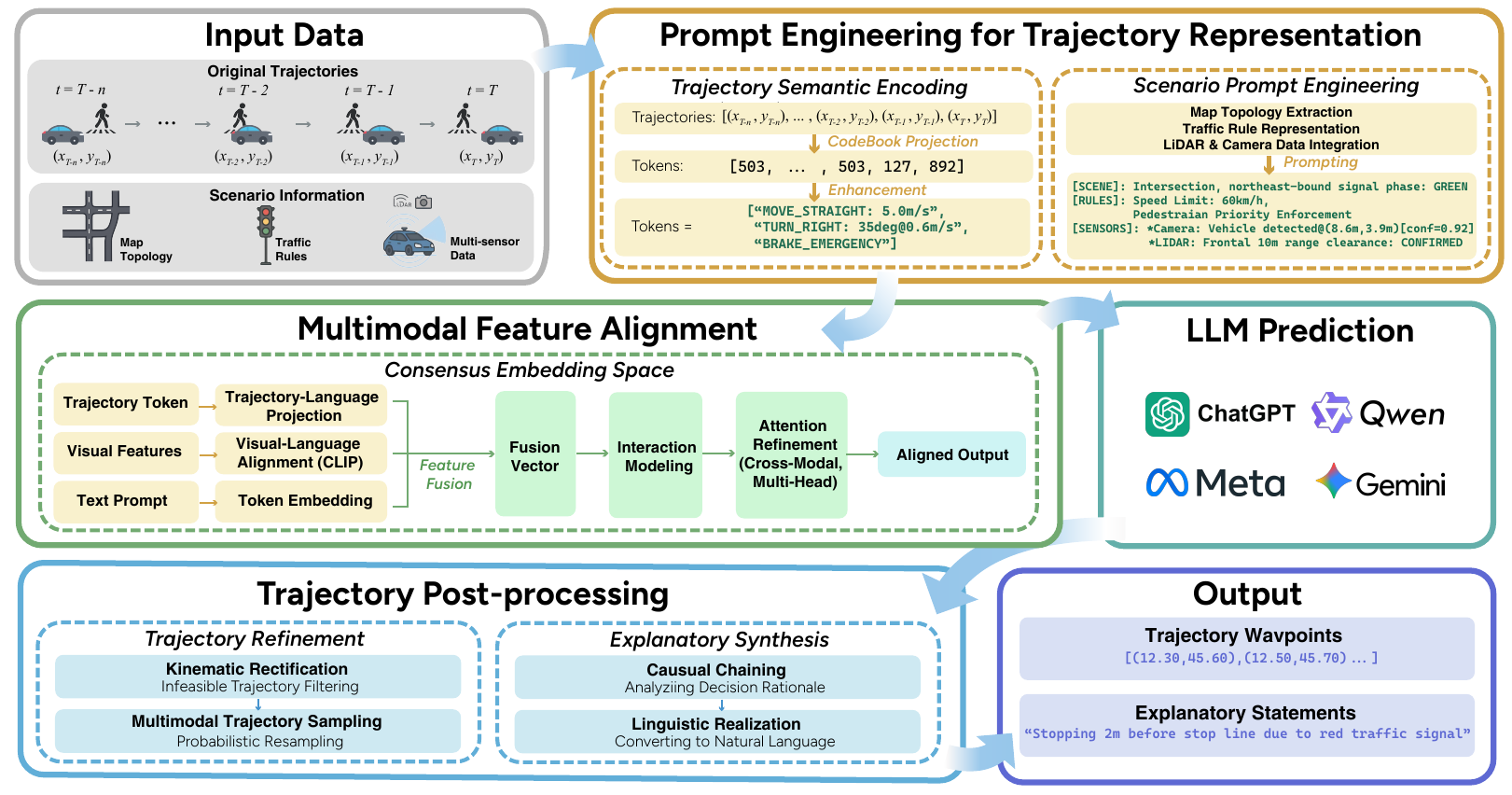}
    \vspace{-4mm}
    \caption{Architectural Overview of LLM-Centric Trajectory Prediction}
    \label{fig:flowchart}
    \vspace{-4mm}
\end{figure*}

\subsection{Trajectory Prediction Formulation}

\subsubsection{Historical States (Input $X$)}
Let $X = \{ s^0, s^1, \dots, s^{t_h-1} \}$ denote the observed states of $N$ traffic participants over the past $t_h$ time steps, where each $s^t = \{ s_1^t, s_2^t, \dots, s_N^t \}$ captures the states of all agents at step $t$. The individual state $s_j^t \in \mathbb{R}^d$ typically includes the 2D position $(x_j^t, y_j^t)$ and can be extended with higher-order dynamics such as velocity, acceleration, heading angle, and other kinematic attributes. The current time corresponds to $t = t_h$.

\subsubsection{Environmental Context (Optional Input $I$)}
To complement agent-level states, contextual information may be available as $I$, including High Definition (HD) maps, road geometry, traffic signals and static obstacles. Formally, $I \subseteq \{\text{HD maps}, \text{traffic lights}, \text{road geometry}, \dots \}$.

\subsubsection{Complete Information (Condition $C$)}
The full input condition for trajectory prediction is defined as the union of dynamic and static information, i.e.,
\[
C = X \cup I.
\]

\subsubsection{Prediction Target (Output $Y$)}
The goal is to predict the future evolution of all agents over the next $t_f$ time steps, denoted as
\[
Y = \{ s^{t_h}, s^{t_h+1}, \dots, s^{t_h + t_f - 1} \},
\]
where each $s^{t_h + k} = \{ s_1^{t_h+k}, \dots, s_N^{t_h+k} \}$ contains the predicted states of all $N$ agents at future step $k$.

\subsubsection{Probabilistic Modeling}
Trajectory prediction is commonly formulated as a conditional probabilistic inference problem, aiming to estimate the posterior distribution
\[
P(Y \mid C) = P(s^{t_h}, \dots, s^{t_h + t_f - 1} \mid X, I),
\]
which captures uncertainty, as well as potential multimodal behaviors inherent in future trajectories.

\subsection{Vehicle Trajectory Prediction}
Vehicle trajectory prediction, which serves as a critical component in the reasoning layer of autonomous driving systems, has evolved from physics rules-based simplified models to data-driven deep learning frameworks. Contemporary approaches achieve high-precision, probabilistic multimodal trajectory prediction over multisecond horizons (typically 3-6 seconds) by integrating historical trajectories, multiagent interaction features, and semantic information from high-definition maps.
\subsubsection{Physics-Based Methods}
These approaches model vehicle trajectories using dynamics or kinematics equations, offering the advantages of computational efficiency and training-free implementation. However, they usually struggle to handle complex interactive scenarios and are generally limited to short-term predictions ($\leq$1 second)  \cite{ammoun2009real, lytrivis2008cooperative}. Representative methods include single-trajectory approaches, which employ simplified models such as Constant Velocity or Constant Acceleration for trajectory prediction, ignoring environmental constraints and interactions\cite{kaempchen2009situation}; Kalman filter-based methods, which characterize state noise via Gaussian distributions and achieve multimodal output through Interacting Multiple Models (IMM) \cite{lefkopoulos2020interaction}; and Monte Carlo techniques, which generate feasible trajectories using stochastic sampling and refine predictions incorporating physical constraints \cite{wang2019trajectory}.

\subsubsection{Machine Learning Methods}
These methods model vehicle behavior patterns using data-driven approaches with predefined driving intentions (e.g., changing lane, following cars), offering strong interpretability, high computational efficiency, and enhanced accuracy for short-term predictions\cite{deo2018would}. However, they exhibit limited generalizability, dependency on manual feature engineering and large-scale annotated data, restricted adaptability to dynamic environments, and difficulties in handling complex traffic interactions.
Representative techniques include Gaussian Processes (GP), which perform probabilistic regression based on a set of trajectory prototypes to capture interactive relationships \cite{tran2014online}; Hidden Markov Models (HMM), which infer implicit driving intentions by treating historical trajectories as observation sequences \cite{deng2018improved, qiao2014self}; and Dynamic Bayesian Networks (DBN), which improve prediction accuracy in complex scenarios through integrated temporal reasoning and interaction modeling \cite{he2019probabilistic}.

\subsubsection{Deep Learning Methods}
Deep learning approaches leverage end-to-end architectures to automatically extract spatiotemporal characteristics, significantly improving long-term prediction accuracy ($\geq$ 5 seconds) and emerging as the current research mainstream\cite{deo2018convolutional}. Deep learning excels in vehicle trajectory prediction by effectively modeling traffic participant interactions through data-driven representation learning, demonstrating robust performance through hierarchical feature fusion in dynamic environments, and generating multimodal probabilistic outputs to capture behavioral uncertainty. These advantages establish deep learning as the prevailing approach in trajectory prediction research. However, significant challenges persist, including prohibitive computational demands, inherent interpretability limitations due to black-box architectures \cite{zablocki2022explainability}, pronounced dependence on extensive annotated datasets, and inadequate uncertainty quantification in safety-critical contexts \cite{li2022uqnet}.    The dominant paradigms comprise sequence models such as RNNs\cite{messaoud2021trajectory}, Long Short-Term Memory Networks (LSTMs)\cite{ding2019online}, and Gated Recurrent Units (GRUs)\cite{tang2019multiple} for temporal dependency modeling, with attention mechanisms (e.g., Transformer) addressing long-range correlations\cite{kim2020multi}. GNNs representing traffic agents and road elements as nodes to encode topological interactions \cite{li2019grip, gao2020vectornet}. Generative models, such as Generative Adversarial Networks (GANs)\cite{kuo2022trajectory} and Conditional Variational Autoencoders (CVAEs)\cite{feng2019vehicle}, have been extensively employed in trajectory prediction due to their capability of generating multimodal trajectories that effectively capture behavioral uncertainty and diversity. These approaches have demonstrated significant improvements in prediction accuracy and robustness through the incorporation of driving knowledge into deep learning models\cite{ding2023incorporating}.

\subsubsection{Reinforcement Learning Methods}
Reinforcement Learning (RL) derives driving policies by learning reward functions from expert demonstrations to generate trajectories balancing safety and operational efficiency \cite{fernando2020deep}. Specifically, these approaches exhibit three principal advantages: effective modeling of interactive scenarios through intention-aware reward structures, robust long-term predictive capability through Markov decision processes, and significant adaptability to novel environments through policy generalization \cite{kuefler2017imitating}. Nevertheless, RL confronts critical limitations including inherent training instability from nonconvex optimization landscapes, stringent dependency on extensive high-fidelity demonstrations (susceptible to distributional bias), challenges in designing verifiable safety-critical reward functions, and limited interpretability of black-box policies impeding formal verification \cite{casper2023open}. Dominant paradigms comprise Maximum Entropy Inverse Reinforcement Learning, which formulates trajectory planning as entropy-regularized multiobjective optimization to maintain behavioral diversity \cite{xu2020learning}; Deep Inverse Reinforcement Learning (DIRL) integrating convolutional architectures to process multimodal sensory inputs (e.g., LiDAR, camera) for unified perception-decision pipelines \cite{zhu2020off}; Generative Adversarial Imitation Learning (GAIL) approximates expert-like policies via minimax adversarial training and demonstrates generalization to states outside the demonstration set\cite{kuefler2017imitating}.

\subsection{Pedestrian Trajectory Prediction}
Pedestrian trajectory prediction is a fundamental perception challenge in autonomous driving systems operating within human-vehicle mixed environments, primarily serving to improve operational safety through anticipatory collision avoidance while ensuring socially aware navigation behaviors \cite{golchoubian2023pedestrian}. 
\subsubsection{Physics-Based Models}
Physics-based approaches employ explicit rules to simulate pedestrian dynamics. The dominant Social Force Model (SFM) calculates motion through attraction forces to destinations and repulsive forces from obstacles and vehicles \cite{chen2018social}, with advanced variants modeling vehicles as ellipses to capture velocity-modulated danger zones \cite{yang2018social}. Complementary methods include kinematic formulations (e.g., constant velocity/acceleration with Kalman filters \cite{zhang2021pedestrian}) and cellular automata using gap acceptance thresholds for interaction decisions \cite{cheng2019study}. These models provide high interpretability and computational efficiency crucial for real-time applications, but suffer from three core limitations: hand-crafted rules cannot adapt to complex social behaviors; deterministic outputs fail to capture trajectory uncertainty; and environment-specific calibration (e.g., using DUT \cite{yang2019top} data) impedes generalization in unstructured spaces \cite{predhumeau2022agent}.

\subsubsection{Data-Driven Methods}
Data-driven methods overcome SFM limitations by learning implicit interaction patterns. Generative models, including CVAEs that sample latent variables for multimodal trajectories and GANs that diversify predictions through noise injection \cite{wang2021multi, wang2021ltn}, employ hybrid frameworks to mitigate mode collapse through structured latent spaces\cite{zhou2022dynamic}. GNNs model heterogeneous interactions through agent nodes and edges encoding relative distance, velocity, and heading angles \cite{li2021interactive, li2021hierarchical}, while hierarchical architectures differentiate pedestrian-vehicle interactions via speed-dependent edge weights\cite{su2022trajectory} and attention mechanisms prioritize critical interactions \cite{girase2021loki}. Despite their capacity to capture nonlinear patterns, these models exhibit three critical limitations: (i) Opacity in decision processes hinders interpretability\cite{cheng2020trajectory}; (ii) Performance depends critically on large-scale annotated datasets, yet such data remain scarce for unstructured environments; (iii) Long-horizon predictions may violate kinematic feasibility, and GANs are susceptible to mode collapse phenomena \cite{gupta2018social}.
\begin{table*}[!h]
    \centering
    \caption{Comparative Analysis of Vehicle Trajectory Prediction Methods}
    \vspace{-2mm}
    \begin{tabularx}{\linewidth}{cp{3cm}XX}
        \toprule
         \textbf{Approaches}   & 
         \textbf{Core Ideas}  &
         \textbf{Advantages} & \textbf{Limitations} \\
        \midrule
        Physics-Based Methods & \tabincell{Governed by kinematic\\dynamic equations and\\ physical constraints} &
        \tabincell{High computational efficiency                     \\ No training required\\  High short-term prediction accuracy \textsuperscript{\ddag}} & \tabincell{Struggles with complex interaction scenarios\\ Limited to long-term predictions \textsuperscript{\#}\\ Ignores environmental constraints\\ }                      \\  \midrule
        Machine Learning Methods     &  \tabincell{Statistical learning with \\pre-defined behavioral\\ intentions}&\tabincell{Strong interpretability                           \\ High computational efficiency\\ Quantifiable uncertainty}                             & \tabincell{Limited generalization ability\\ Weak interaction modeling capability\\ Difficulty in modeling complex distributions}                                   \\ \midrule

        Deep Learning Methods     &  \tabincell{End-to-end learning of \\complex spatio-temporal \\representations from data} & \tabincell{Improved long-term porediction 
        accuracy  \\ Effective interaction modeling\\ Multimodal uncertainty capture}                             & \tabincell{Prohibitive computational demands\\Inherent interpretability limitations                      \\ Extensive annotated datasets dependence}                                   \\ \midrule
        
        Reinforcement Learning Methods& \tabincell{Derive driving policy by\\ learning reward functions\\from expert demonstrations}&\tabincell{ Novel environment adaptability\\ Interactive scenario modeling\\ long-term prediction robustness}                    & \tabincell{Training is complex and unstable\\ High-fidelity data dependence\\Black-box interpretability limitations\\ Safety reward design difficulty\\ } \\
        \bottomrule
    \end{tabularx}
    \vspace{0.1em}
    \begin{flushleft}
    \small
    \textsuperscript{\ddag} “short-term prediction” refers to forecast with a time horizon of $\leq 1$ second.\\
    \textsuperscript{\#} "long-term prediction" refers to forecast with a time horizon of $\geq$ 5 seconds.\\
    \end{flushleft}
    \vspace{-4mm}
\end{table*}

\subsubsection{Hybrid Approaches}
Hybrid frameworks synergistically combine physical priors with data-driven learning. The expert-data fusion strategy overrides implausible deep learning outputs with SFM or game-theoretic conflict resolution\cite{johora2020agent}, while physics-guided learning enforces dynamic constraints through kinematic layers \cite{li2021spatio}. This dual methodology improves robustness by marrying data-driven accuracy with physics-based safety, particularly in shared spaces. However, it introduces implementation complexity, requires meticulous environment calibration, and lacks theoretical convergence guarantees\cite{predhumeau2022agent}.

\section{Trajectory Prediction with LLMs}

Building upon the historical context established in Section II, this chapter delineates a paradigm shift from conventional pattern recognition to semantic-aware cognitive reasoning for trajectory prediction. 
This shift is reflected in the rapidly growing body of academic research, as shown in Fig.~\ref{fig:publications}, which charts the publications distribution in top-tier venues from 2023 to 2025 September. As illustrated in Fig.~\ref{fig:flowchart}, the proposed LLM-Centric Prediction Framework introduces a unified architecture comprising three synergistic pillars:

Multimodal Alignment: Heterogeneous inputs (kinematic trajectories, visual scenes, and linguistic prompts) undergo domain-specific encoding (e.g., VQ-VAE discretization for trajectories, CLIP for visual-language grounding) to generate dimensionally consistent embeddings;

Consensus Embedding Fusion: Cross-modal attention mechanisms refine these embeddings into a unified semantic space, enabling joint representation of spatio-temporal dynamics and contextual semantics;

Constraint-Guided Reasoning: CoT decomposition integrates traffic rules and physical constraints (e.g., ``yield to pedestrians at crosswalks") through autoregressive language modeling, generating both geometrically precise trajectories and natural-language rationales.
This framework fundamentally addresses the interpretability gap in black-box deep learning approaches while enhancing generalization in long-tail scenarios.

\subsection{Perception and Scene Understanding}
Enhancing the comprehension of complex scenarios enables holistic perception of the environment, forming the foundation for accurate behavior prediction of traffic participants and improved safety.
\subsubsection{Vision-Language Models (VLMs) for Object Detection}

\paragraph{Technical Integration} VLMs advance autonomous driving perception by integrating visual sensing with semantic understanding, achieving open vocabulary detection, multimodal conditional detection, and long-tail scenario generalization \cite{xu2024vlm,tian2024drivevlm}. Technically, they combine complementary strengths of traditional perception systems and language models. For example, DriveVLM-Dual integrates VLM-based semantic scene analysis with conventional 3D object detection (e.g., IoU-based cross modal alignment), using linguistic descriptions to enhance recognition of rare objects (e.g., road debris and irregular vehicles) while leveraging geometric detectors for precise 3D localization, thus effectively mitigating the spatial reasoning limitations of VLMs \cite{tian2024drivevlm}.

\paragraph{Role in Explainability} VLMs generate human interpretable object-level descriptions, serving as intermediate representations for decision-making. Traditional perception systems (e.g., object detection, semantic segmentation) provide structured environmental data, but their outputs lack intuitive explainability. VLMs bridge this gap by converting vectorized object states (e.g., position, velocity) into natural language, enabling contextual reasoning about detected entities\cite{chen2024driving}.

\paragraph{Synergistic Integration} The language-augmented detection output of VLMs directly serves planning modules. RAG-Driver further introduces Retrieval-Augmented In-Context Learning (RA-ICL), dynamically retrieving expert demonstrations from similar scenarios to enhance zero-shot generalization (e.g., unseen London roads) while jointly optimizing control signals and natural language explanations\cite{yuan2024rag}.

\begin{table*}[!h]
    \centering
    \caption{Comparative Analysis of Pedestrians Trajectory Prediction Methods}
    \vspace{-2mm}
    \begin{tabularx}{\linewidth}{cp{3.3cm}p{3.6cm}p{4.3cm}p{4cm}}
        \toprule
        \textbf{Approaches}  & 
        \textbf{Representative Models}&
        \textbf{Advantages} & 
        \textbf{Limitations} &
        \textbf{Applicable Scenarios} \\
        \midrule
        Physics-Driven Methods & 
        \tabincell{Social Force Model\\Constant Velocity/Acceleration\\Kalman Filter + IMM}&
        \tabincell{High interpretability                    \\ Computational efficiency\\ Training-free deployment} & 
        
        \tabincell{  Limited behavioral flexibility\\  Poor generalization to new environment\\  Homogeneous predictions}  &
        \tabincell{Structured \& Shared-space\\Low interaction density\\Short-term prediction  \textsuperscript{\ddag}}
        \\  \midrule
        Data-Driven Methods     & 
        \tabincell{LSTM/GAN/CVAE\\Graph Neural Networks \\Transformer-based Models\\Reinforcement Learning}&
        \tabincell{Implicit complex pattern learning                       \\Multi-modal trajectory generation\\ High short-term accuracy}                             & \tabincell{Low interpretability\\ Data dependency and bias propagation\\ Long-term kinematic infeasibility risks\textsuperscript{\#}}   &
        \tabincell{High interaction density\\Complex environments\\Long-term prediction }\\ \midrule
        Hybrid Approaches  & 
        \tabincell{Game-Theoretic SFM\\ SFM/GT-Augmented DL\\Physics-Guided Learning}
        &\tabincell{  Physically feasible trajectories\\ Explicit conflict resolution\\ Socially compliant}      & \tabincell{Implementation complexity\\ Calibration dependency\\ No convergence guarantees\\ }
        &
        \tabincell{Mixed traffic environments\\Unstructured environments\\Interaction scenarios }
        \\
        \bottomrule
    \end{tabularx}
    \vspace{0.1em}
    \begin{flushleft}
    \small
    \textsuperscript{\ddag} “short-term prediction” refers to forecast with a time horizon of $\leq 1$ second.\\
    \textsuperscript{\#} "long-term prediction" refers to forecast with a time horizon of $\geq$ 5 seconds.\\
    \end{flushleft}
    \vspace{-4mm}
\end{table*}

\subsubsection{Scenario Description Generation}
\paragraph{Multimodal Perception}
Autonomous vehicles integrate heterogeneous sensors (e.g., cameras, LiDAR, millimeter-wave radar), each with distinct advantages and limitations in environmental perception \cite{yurtsever2020survey, sima2024drivelm}. MLLMs demonstrate significant capabilities in interpreting non-textual data (images, point clouds) through textual analysis \cite{zhong2024language}, with foundational frameworks like CLIP generating image representations via image-text contrastive learning \cite{radford2021learning} and LLaMA integrating visual encoders with LLMs to enhance joint understanding of visual-linguistic concepts \cite{liu2024improved}. Domain specific advancements include: DriveGPT4, which converts video inputs into driving relevant textual responses\cite{xu2024drivegpt4}; HiLM-D, which incorporates high resolution visual details to improve hazard recognition\cite{ding2025hilm}; Talk2BEV, which fuses Bird's-Eye View (BEV) representations with linguistic context using pretrained vision language models \cite{choudhary2024talk2bev}; and Da Yu, an efficient MLLMs that fuses RGB camera streams with language modeling via a Nano Transformer Adaptor, generating long text navigational captions through spatiotemporal fusion \cite{guan2025yu}.

\paragraph{Sensor Fusion Paradigm}
LiDAR point clouds processed by 3D object detectors yield scene-level feature maps and object-level vectors\cite{rong2023dynstatf}. When occlusion compromises a cooperative autonomous vehicle's sensor visibility (e.g., due to large obstructions such as trucks or buildings), centralized multi-agent perception systems leverage feature fusion from connected vehicles and infrastructure to resolve blind spots. For example, heterogeneous graph-based transformers dynamically weight infrastructure features in occluded regions, significantly improving detection robustness\cite{xu2022v2x}. Subsequent safety-critical advisories(such as trajectory adjustment directives) are generated by multi-modal LLMs integrating these fused perceptions, exemplified by responses like: ``Vehicle detected behind reference object; adjust trajectory to avoid collision" \cite{chiu2025v2v}. This multimodal fusion enables holistic decision-making by contextualizing visual data with semantic cues \cite{shao2024lmdrive}.

\paragraph{Generative Scene Modeling}Models like GAIA-1 synthesize realistic driving scenarios from video, text, and action input, predicting potential outcomes of dynamic interactions \cite{hu2023gaia}. VisionTrap proposes a generative scene modeling framework that integrates surround-view camera inputs with generative textual supervision, derived from VLM/LLM-crafted descriptions, to semantically enrich BEV representations \cite{moon2024visiontrap}. DriveDreamer-2 \cite{zhao2025drivedreamer}, where an LLM interprets natural language queries (e.g., ``simulate a highway cut-in with aggressive braking") to generate agent trajectories and HD maps, feeding a diffusion-based world model that renders spatiotemporally consistent multi-view videos.

\paragraph{Scene Semantic Parsing}
LLMs elevate environmental perception to semantic cognition through multimodal alignment and causal reasoning capabilities \cite{wang2023drivemlm, yang2025multimodal}. Exemplified by frameworks like DualDiff+, which Semantic Fusion Attention (SFA) dynamically integrates multimodal inputs to improve complex scenario understanding \cite{yang2025dualdiff+}. 
LINGO-1 pioneers scene semantic parsing by unifying sequential visual features and natural language into a joint embedding space via CLIP-style contrastive learning. This approach transcends conventional geometric primitives (e.g., bounding boxes) to generate structured scene semantics, exemplified by \textit{``Pedestrians crossing at marked crosswalk"} (\textit{spatial-functional parsing}) and \textit{``Parked van partially obstructing driving lane"} (\textit{object-state-context parsing}). These natural language descriptors transform raw pixels into actionable scene graphs for downstream prediction and planning modules\cite{WayveLINGO2023}.

\subsection{LLM-based Vehicle Trajectory Prediction}
The evolution of LLM-based trajectory prediction methods has established a tripartite knowledge hierarchy, which spans representational mapping, multimodal fusion, and constrained reasoning, that extends beyond conventional pattern recognition by incorporating structured semantic representations\cite{keysan2023can}. 

\subsubsection{Trajectory-Language Mapping Methodologies}
\paragraph{Prompt Design for Trajectory Modeling}
Prompt engineering has emerged as a pivotal methodology for mediating multimodal inputs and trajectory generation in LLMs, evolving beyond rudimentary instruction crafting into sophisticated knowledge-guided frameworks. Contemporary approaches construct structured textual prompts that encode traffic scene semantics (e.g., road topology, traffic regulations) and agent dynamics (e.g., position, velocity) to elicit scene embeddings for trajectory inference \cite{keysan2023can}. STG-LLM proposes a tokenization strategy where each node in the spatial graph is treated as a token, encapsulating its time-series observations and temporal semantics (e.g., time-of-day embeddings). This avoids the inefficiency of natural language descriptions and enables LLMs to capture spatio-temporal dependencies via multihead attention mechanisms\cite{liu2024can}. S4-Driver processes historical ego-states (positions, velocities, accelerations) 
as floating-point values within text prompts, while high-level behavioral commands (e.g., ``Turn left") are provided as natural language instructions. This unified textual representation avoids explicit kinematic parameterization 
and enables seamless integration with the multimodal encoder\cite{xie2025s4}, while iMotion-LLM employs structured templates with system-defined roles (e.g., specifying coordinate systems and output formats) and user-provided real-time observations (e.g., traffic signs, agent states) to facilitate trajectory prediction and instruction feasibility assessment\cite{felemban2024imotion, peng2025lc}. CoT reasoning further augments this framework through multi-step prompts that integrate contextual background, interaction analysis, and risk assessment to generate structured semantic annotations \cite{liao2025cot}. Such prompts frequently incorporate intent instructions (e.g., ``Ego vehicle turn left"), scene descriptors (e.g., traffic lights, pedestrian behaviors) and spatial directives (e.g., ``Yield to car at specified locations") \cite{xing2025openemma}, enabling nuanced environment-response alignment. Concurrently, multimodal QA pairs operationalize interactive prompting by fusing visual tokens (e.g., BEV embeddings) with domain-specific queries (scene perception, behavior forecasting) to steer joint reasoning \cite{hegdeDistillingMultimodalLarge2025a}, as demonstrated in VQA frameworks that transform kinematic interactions (e.g., ``Lead vehicle: 8.1 m/s; ego vehicle: 7.6 m/s; gap: 11.8 m" \cite{chen2024genfollower}) into actionable trajectory predictions. Collectively, these strategies establish a tripartite knowledge hierarchy: physical (numerical/linguistic encoding of kinematics \cite{liu2024can, xie2025s4}), semantic (logic chains binding scenes to rules \cite{liao2025cot, xing2025openemma}), and evolutionary (self-optimizing prompt iterations ) thereby enhancing the explainability, adaptability, and safety compliance of trajectory prediction systems.
\paragraph{Trajectory Discretization and Vocabulary Construction}
Trajectory discretization bridges continuous motion dynamics with discrete language representations through structured tokenization schemes. Bézier curve encoding converts lane geometries into fixed control points, constructing dedicated vocabularies for spatial context representation \cite{keysan2023can}. Driving trajectories undergo explicit discretization via VQ-VAE, where front-view image frames are tokenized into discrete visual tokens, and frame-wise relative motions are quantized into action bins\cite{chen2024drivinggpt}. Similarly, car-following behaviors in GenFollower are segmented into timestep-based state descriptors (e.g., velocity, spacing), forming structured natural language prompts compatible with LLMs\cite{chen2024genfollower}. Spatiotemporal graph tokenizers (e.g., STG-LLM) decompose complex dynamics into node-level tokens encapsulating time series data and temporal semantics \cite{liu2024can}, while iMotion-LLM discretizes trajectories into directional/speed/acceleration sequences for textual conversion \cite{felemban2024imotion}. Multiview scene encoders further advance this paradigm by generating high-dimensional tokens representing environmental maps, ego vehicles, and agent dynamics, effectively structuring scene components into learnable vocabularies \cite{hegdeDistillingMultimodalLarge2025a}. Causal language modeling frameworks formalize this process, representing trajectories as discrete motion token sequences trained with autoregressive losses for multi-agent prediction \cite{peng2025lc, seff2023motionlm}. In contrast, some approaches eschew discretization; for example, S4-Driver maintains geometric continuity by directly regressing floating-point trajectory coordinates, highlighting a fundamental methodological divergence from token-based methods\cite{xie2025s4}. Collectively, these techniques establish a spectrum from granular discretization to semantic abstraction, balancing representational fidelity with computational tractability.

\subsubsection{Multimodal Fusion Architectures}
\paragraph{Unified Encoder Design for Vision-Language-Trajectory Alignment}
A key innovation driving recent progress in LLM-based trajectory prediction is the development of unified encoder architectures designed to achieve effective alignment of vision-language trajectory. These systems typically employ specialized modules: a language instruction encoder extracts semantic features from textual prompts (e.g., navigation commands, behavioral descriptions) ; an interaction-aware encoder captures complex spatial dynamics and dependencies between the target agent and surrounding entities; and a cross-modal encoder integrates and refines these semantic and spatial features into a coherent representation for prediction \cite{liao2025cot}. Crucially, several frameworks leverage shared scene encoders to establish a common grounding point. For example, DiMA\cite{hegdeDistillingMultimodalLarge2025a} and related visual foundation planners share a scene encoder that converts visual inputs into structured BEV token embeddings, which serve as the fundamental fusion point for visual, linguistic and trajectory information fed into both the planning transformer and the LFMs, allowing structured visual exploitation by the language model \cite{xu2024drivegpt4}. Similarly, DrivingGPT utilizes a pre-trained visual encoder to transform front-view visual inputs into discrete embeddings, coupled with action quantization for trajectory signals and a language tokenizer. These modalities are unified into an interleaved token sequence, allowing multimodal alignment through autoregressive prediction of the next token\cite{chen2024drivinggpt}. This paradigm extends to architectures that feature different visual and language encoders, where components such as Q-Former explicitly align extracted visual features with text tokens, ensuring effective cross-modal fusion for unified representation \cite{liu2025dsdrive}. Further designs focus on domain-specific fusion, iMotion-LLM encodes complex scene data into embeddings projected into the LLM space alongside embedded textual instructions, and this design achieves explicit alignment between vision-inspired vector data, language instructions, and trajectory outputs, forming a cohesive framework for interactive motion planning\cite{felemban2024imotion}. An alternative, highly integrated approach reformulates the prediction task itself as a Visual Question Answering (VQA) problem within a pre-trained LLM like Gemini, where all non-sensor inputs and outputs are represented as pure text via task-specific prompting, leveraging the model's inherent world knowledge while bypassing explicit custom encoder design for sensor fusion \cite{xing2025openemma}. Collectively, these diverse convergent encoder strategies, ranging from modular multi-stream processing with explicit alignment to shared latent space projection and task reformulation, provide the critical infrastructure for aligning heterogeneous vision, language, and trajectory modalities within LLMs, enabling robust and context-aware future motion prediction. 

\paragraph{Interaction Modeling with LLMs}
Recent advancements leverage the intrinsic reasoning capabilities of LLMs to model complex agent interactions in trajectory prediction, transcending traditional pattern recognition through cognitive simulation and structured representation learning. CoT prompting enables LLMs to decompose complex interactions into sequential reasoning steps, such as intent recognition, risk assessment, and maneuver prediction, thus allowing systems like CoT-Drive to interpret agent behaviors (e.g., predicting lane changes by analyzing relative speed to leading vehicles) through explicit modeling of agent-environment dependencies\cite{liao2025cot}. This paradigm extends to tokenization of spatio-temporal interactions. DrivingGPT formalizes driving scenarios as a multimodal driving language with interleaved image action tokens, utilizing autoregressive transformers to model temporal dynamics across frames\cite{chen2024drivinggpt}, while STG-LLM converts Spatial-Temporal Graph (STG) data into discrete tokens via its STG tokenizer, encapsulating the temporal dynamics per graph node and implicitly representing spatial semantics through token relationships. The STG-Adapter then decodes these representations for prediction based on LLM-processed features\cite{liu2024can}. Further innovations formalize interaction as language modeling tasks. GenFollower encodes car-following dynamic states as structured prompts, leveraging LLMs to predict longitudinal vehicle interactions through natural language reasoning chains\cite{chen2024genfollower}. MotionLM discretizes continuous trajectories into motion tokens and employs autoregressive decoding to generate joint distributions of interactive agents, preserving multi-agent coordination in predictions \cite{seff2023motionlm}. To enhance geometric grounding, hybrid architectures integrate vision experts: S4-Driver aggregates multi-view, multi-frame sensor data into 3D sparse volume representations, enriching LLM inputs with structural cues for spatial relationship reasoning \cite{xie2025s4}; openEMMA integrates YOLO-based object detection with MLLMs to contextualize agent states (position, velocity) against scene semantics, enabling interaction-aware trajectory prediction \cite{xing2025openemma}. Simultaneously, knowledge distillation frameworks optimize interaction modeling efficiency, where compact LLMs trained via dual-head coordination modules synchronize reasoning and planning objectives, distilling interaction patterns from complex datasets\cite{liu2025dsdrive}. Crucially, methodologies like iMotion-LLM demonstrate that textual instruction tuning (e.g., using InstructWaymo) aligns LLM priors with interactive scenario features, generating socially compliant trajectories conditioned on linguistic constraints \cite{felemban2024imotion}. Collectively, these approaches harness LLMs not merely as predictors but as interactive models, by transforming raw sensor data into explainable, structured representations of agent dynamics, and fundamentally advancing the fidelity of socially aware trajectory forecasting.

\begin{table*}
\centering
\caption{Breakthroughs in hybrid prediction paradigms}
\vspace{-2mm}
\label{tab:hybrid}
\begin{tabular}{lccc}
\toprule
\textbf{Architecture} & \textbf{Core Innovation} & \textbf{Performance Gain} & \textbf{Scenario} \\ 
\midrule
TCP Framework & Dynamic fusion of trajectory GRU and control ROACH & 37\% reduction in collision rate & CARLA simulations (sharp turns) \\
DriveSuprim & Rotation enhancement + self-distillation soft labels & SOTA 93.5\% PDMS (NAVSIM v1) & Extreme steering conditions \\
AutoDRRT 2.0 & BEVFormer quantization + sparse compression & $<$100ms end-to-end latency & Multi-sensor fusion \\
\bottomrule
\end{tabular}
\vspace{-4mm}
\end{table*}

\subsubsection{Reasoning-based Prediction Frameworks}
\paragraph{Commonsense Reasoning and Causal Trajectory Modeling}
The integration of commonsense knowledge and causal dynamics into trajectory prediction represents a paradigm shift where LLMs transcend pattern recognition to emulate human-like contextual reasoning. This approach leverages LLMs as repositories of domain-specific priors (e.g., traffic conventions, safety margins) and causal mechanisms (e.g., acceleration as a response to obstacles), fundamentally enhancing prediction robustness. Explicit knowledge injection occurs through semantic grounding: pre-trained language encoders extract scene semantics containing normative behavioral patterns (e.g., yielding protocols at intersections), enabling predictions aligned with traffic conventions \cite{keysan2023can}. Similarly, TOKEN utilizes object-level semantics (traffic cones, vehicle states) and structured reasoning with traffic rules to generate causally consistent trajectories through LLM-based rule-guided trajectory generation\cite{tian2024tokenize}. Concurrently, based on the rigorous CoT prompting framework, CoT-Drive leverages object-level semantics and embedded traffic rules to first generate contextual scene descriptions, then progressively refines its reasoning through four structured phases: statistical scene analysis, multi-agent interaction assessment, collision risk quantification, and finally trajectory prediction with explicit commonsense compliance \cite{liao2025cot}. GenFollower integrates comfort/safety constraints directly into prompts, requiring stepwise justification for car-following decisions \cite{chen2024genfollower}. LC-LLM reframes lane changes as language modeling tasks with CoT supervision, generating explainable predictions rooted in causal antecedents (e.g., ``gap acceptance due to approaching highway exit") \cite{peng2025lc}. To democratize reasoning capabilities, knowledge distillation transfers commonsense reasoning from VLMs to compact LLMs via structured CoT datasets, enforcing task consistency in resource-constrained systems \cite{liu2025dsdrive}. Collectively, these methodologies position LLMs as structured reasoning engines, transforming raw sensor data into rule-governed prediction that encode traffic conventions and safety constraints. By integrating commonsense priors via semantic grounding and multi-step causal reasoning they enhance robustness beyond purely statistical approaches, though counterfactual validity requires further verification.
\paragraph{Linguistic Constraints for Traffic Rule Compliance}
The enforcement of traffic norm compliance in LLM-based trajectory prediction has been revolutionized through formalized linguistic constraints, where natural language directives explicitly encode regulatory principles and safety protocols to mitigate distributional shift between training data and real-world scenarios. This paradigm operationalizes traffic rules as action-guiding prompts, instructional conditions, and structured reasoning frameworks that intrinsically constrain model outputs. Systems like DriveGPT4 leverage CoT prompting to decompose rule adherence into verifiable reasoning steps requiring explicit justification for signal compliance (e.g., ``decelerate due to red light") and right-of-way conventions \cite{xu2024drivegpt4}. Similarly, conditioned instruction architectures embed regulatory semantics directly into inputs. iMotion-LLM rejects infeasible maneuvers by validating textual commands (e.g., ``turn left") against scene context \cite{felemban2024imotion}.  TOKEN integrates road-level navigation commands (e.g., ``turn right at the intersection'') with object-centric tokens (e.g., encoding stop signs) to ensure navigational compliance via structured CoT reasoning, reducing collisions by 39\% in long-tail scenarios\cite{tian2024tokenize}. Crucially, explicit safety primitives are engineered into system prompts, where GenFollower hardcodes minimum following distances and comfort thresholds as inviolable linguistic boundaries \cite{chen2024genfollower}. S4-Driver encodes turn commands (e.g., ``left"/``right") as high-level behavior inputs to guide planning trajectories. This leverages linguistic constraints for traffic rule compliance at the navigational level, while low-level rules are implicitly learned from data\cite{xie2025s4}. Collectively, these linguistic strategies operationalize traffic regulations as executable cognitive frameworks, enabling LLM-based planners to generate trajectories that adhere to navigational rules through explicit constraints while implicitly learning safety-critical behaviors from data, thus enhancing functional compliance in real-world distribution shifts.
\begin{table*}[h]
\centering
\caption{Road safety performance of trajectory prediction systems}
\vspace{-2mm}
\label{tab:safety}
\begin{tabular}{lccc}
\toprule
\textbf{Manufacturer} & \textbf{Collision Rate (/M miles)} & \textbf{Certification} & \textbf{Key Metric} \\ 
\midrule
Waymo\cite{waymo2023safety,favaro2023building} & 0.08 & ISO 21448 & 92\% reduction in pedestrian injury rate \\
Tesla Autopilot\cite{TeslaVehicleSafetyReport} & 0.15 &  NHTSA  5-stars & 87\% reduction in vehicle fire probability \\

\bottomrule
\end{tabular}
\vspace{-4mm}
\end{table*}
\subsubsection{LLMs for End-to-End Trajectory Prediction}
\paragraph{Language-Driven End-to-End Cognitive Prediction}
Traditional end-to-end autonomous driving frameworks exhibit critical limitations in capturing underlying cognitive processes (e.g., understanding driver intent, anticipating interactions, reasoning about scene context), lacking genuine comprehension of driving tasks and interpretability. These systems often operate as black boxes, failing to effectively ground semantic context (e.g., ``aggressive cut-in intent") in the generation of trajectory predictions\cite{winter2025generative, chen2024end}. LLMs offer transformative potential by leveraging their dual capabilities: contextual reasoning for scene interpretation and language generation for step-by-step rationale provision\cite{winter2025generative}. However, integrating LLMs into real-time driving systems faces fundamental tensions between computational efficiency and the unsolved challenge of mapping high-level textual reasoning to precise trajectory coordinates\cite{winter2025generative}.
To address these tensions, recent research explores multimodal fusion and language-centric paradigms. One pioneering work \cite{keysan2023can} encoded scene geometry into text using Bézier curves, reducing the size of the lane geometry representation by 56\%. Parallel frameworks like CoT-Drive decompose predictions into structured reasoning chains via CoT prompting: first, identifying conflict actors (e.g., ``pedestrian waiting to cross''), then inferring behavioral constraints (e.g., ``deceleration likely''), and finally guiding a trajectory decoder to output compliant coordinates\cite{liao2025cot}. Senna adopts a similar language-centric approach but decouples high-level meta-action generation (for example, ``turn left") and low-level trajectory prediction, where a vision-language model reasons over the scene and outputs interpretable decisions, which are translated into precise coordinates by a downstream end-to-end planner\cite{jiang2024senna}. DSDrive achieves cognitive decision making through language driven explicit CoT reasoning and directly outputs interpretable trajectory predictions, seamlessly integrating high-level semantic understanding with low-level motion planning within a lightweight, unified system\cite{liu2025dsdrive}. Closed-loop architectures represent a third evolutionary pathway. LMDrive pioneers this paradigm by integrating LLMs into a closed-loop framework, where natural language instructions dynamically guide real-time control predictions in response to environmental feedback\cite{shao2024lmdrive}.

\paragraph{End-to-End Trajectory Prediction in Automotive Systems}
Leading automakers are actively exploring end-to-end trajectory prediction systems that increasingly incorporate LLMs, moving away from conventional modular pipelines toward integrated cognitive architectures. This transition is not merely technical but reflects a broader shift toward systems capable of more interpretable and context-aware environmental understanding. The integration of LLMs has contributed to improvements in reasoning-based prediction performance while introducing new challenges for safety certification and validation methodologies. Furthermore, hardware-algorithm co-design is crucial in achieving low prediction latency, making these models viable on leading automotive computing platforms. Tesla's 2024 Impact Report \cite{tesla2024impact} reports that vehicles with Autopilot enabled experience one accident per 6.77 million miles. This safety record is enabled by advances in low-latency perception and prediction required for real-time decisions. The performance of the system is further demonstrated in challenging real-world scenarios, such as performing unprotected left turns in dense traffic \cite{TeslaOracleCookFSD2024}. Waymo's approach integrates the MotionLM architecture, which uses discrete tokenization and autoregressive decoding for efficient trajectory forecasting\cite{seff2023motionlm}. This synergy enables real-time performance, robust perception, and reliable prediction of complex interactions, making advanced models viable in automotive computing systems. 
Li Auto HaloOS reduces the IMU-to-trajectory latency below 1 ms by deterministic sensor fusion \cite{li2025haloos}. Xiaomi's HAD architecture has been reported to utilize a new road foundation model built on BEV frameworks to achieve low-latency image-to-trajectory conversion\cite{xiaomipilot2024}. Certification benchmarks confirm an 8-fold reduction in collision probability for these systems compared to industry averages, mainly attributable to improved temporal prediction accuracy \cite{tesla2024impact,mobileye2025eyeq6}.

\begin{table*}[!h]
    \centering
    \caption{Comparative analysis of LLM-based paradigms for pedestrian trajectory prediction}
    \vspace{-2mm}
    \label{tab:llm pedestrians}
    \begin{tabularx}{\linewidth}{cXXX}
        \toprule
        \textbf{Category}  & 
        \textbf{Core Idea}&
        \textbf{Advantages} & 
        \textbf{Limitations} 
        \\
        \midrule
        Language-Based Prediction & 
        \tabincell{Uses LLMs to extract motion cues\\ from history, enhanced by clustering}&
        \tabincell{Semantic motion representations\\ Uncertainty quantification capability} & \tabincell{High prompt sensitivity\\
        Substantial computational overhead\\ Extensive pre-processing required}
        \\  \midrule
        LLM-Driven Evolution     & 
        \tabincell{LLM + evolutionary algorithms to \\auto-design prediction heuristics}&
        \tabincell{Full automation\\ Interpretable rule generation\\ Exceptional cross-dataset generalization}  &
        \tabincell{May underperform on single datasets\\ Neglects contextual cues\\Lack task-specific optimization}\\ \midrule
        LLM Agent Simulation  & 
        \tabincell{LLMs generate semantic activities\\Physical models handle space.}
        &\tabincell{Behaviorally plausible trajectories\\ High interpretability\\ Reduced real-data dependency } & \tabincell{Limited physical fidelity\\ High computational cost\\Contextual scene information}
        \\ \midrule
        Prompt-Based Prediction &
        \tabincell{Uses VLM for zero-shot behavior \\prediction} &
        \tabincell{Training-free deployment\\ Holistic scene understanding\\ Group behavior reasoning} &
        \tabincell{Image quality sensitivity\\ Poor relative motion modeling\\ Output inconsistency issues}
        \\ \midrule
        CoT Visual Reasoning &
        \tabincell{Visual prompts + Chain-of-Thought \\to boost goal/trajectory prediction} &
        \tabincell{High goal prediction accuracy\\ User-controllable generation\\ Transparent reasoning pathway} &
        \tabincell{Sensitivity to prompt design\\ Increase computational overhead\\ Context-dependent performance}
        \\
        \bottomrule
    \end{tabularx}
    \vspace{-4mm}
\end{table*}

Hybrid prediction paradigms represent a critical advance in handling diverse driving scenarios. As detailed in Table \ref{tab:hybrid}, the TCP framework dynamically integrates trajectory GRU networks with ROACH control expert models, reducing collision rates by 37\% in CARLA simulations during sharp-turn scenarios\cite{wu2022trajectory}. DriveSuprim employs rotation enhancement and self-distillation soft labels, achieving a state-of-the-art score of 93.5\% PDMS in NAVSIM v1. Its performance gains were particularly pronounced in extreme steering scenarios, outperforming prior models by up to 2.9\% on sharp-turn subsets of the benchmark\cite{yao2025drivesuprim}. Meanwhile, AutoDRRT 2.0 achieves sub-100 ms end-to-end latency through a combination of techniques including BEV model quantization, structured sparsity, operator optimization, and distributed computing\cite{AutoDRRT2024}. These innovations demonstrate significant performance gains across varied operational contexts, addressing long-standing challenges in multi-agent interaction modeling.

The genetic validation and safety certification frameworks have undergone transformative development. Wayve's GAIA-1 synthesizes driving scenarios at 288×512 resolution using discrete token sequences, demonstrating the potential to generate large quantities of edge cases that could substantially reduce the reliance on real world test miles\cite{wayve2023gaia1,hu2023gaia}. Waymo’s comprehensive safety pyramid integrates three validation tiers: extensive simulated miles incorporating numerous edge scenarios; closed-course testing facilities; and millions of autonomous miles in the real world in urban environments. This integrated approach achieves a 92\% reduction in pedestrian collision rates, as quantified in Table\ref{tab:safety}\cite{waymo2023safety}. The NIO Banyan system leverages a generative AI-powered world model to simulate and anticipate critical driving scenarios, significantly improving decision making and safety \cite{niosmarttech2024}. This paradigm transition to simulation-based validation is essential to manage the increasing cost and complexity of verification and validation for automotive software, which is projected to constitute 29\% of the total software market by 2030. It broadens the coverage of low-probability safety-critical scenarios that are prohibitively expensive or dangerous to physically test \cite{mcKinsey2024}.

\paragraph{Convergence Trends and Future Directions}
Despite these advances, significant challenges persist. Long-tail generalization is severely impeded by heavy precipitation, as it critically degradesthe integrity of the LiDAR point cloud, leading to substantially amplified prediction errors in downstream tasks\cite{dreissig2023survey}. Human-vehicle interaction scenarios, especially unprotected left turns, represent a high-risk maneuver where automated vehicles exhibit a quantifiable probability of collision, as demonstrated in dedicated safety validation studies\cite{corso2020scalable}. SOLVE epitomizes the convergence trend of VLMs and end-to-end autonomous driving networks, achieving synergistic integration through feature-level knowledge sharing (via SQ-Former) and trajectory-level cooperation (via T-CoT and temporal decoupling), marking future directions for robust trajectory prediction via multimodal fusion and efficient reasoning paradigms\cite{chen2025solve}. OpenDriveVLA's scalable VLA architecture, which offers models of varying complexity (e.g., 0.5B to 7B parameters), demonstrates the potential for hardware-algorithm co-design in autonomous driving. By achieving competitive performance even with its smallest model, it provides a viable path to dynamic computational resource allocation, suggesting that future real-time systems could adjust model complexity to optimize hardware utilization\cite{zhou2025opendrivevla}.
The convergence of physics-informed learning (exemplified by Waymo's MotionLM\cite{seff2023motionlm}), generative validation (Wayve's GAIA\cite{wayve2023gaia1}), and hardware-aware compression (AutoDRRT 2.0\cite{AutoDRRT2024}) is steering trajectory prediction toward verifiable safety and economical deployment. As Trent Victor, Waymo's Director of Safety Research and Best Practices, posits: ``This research adds to the growing body of scientific evidence regarding the life-saving potential of the Vision Zero principles and helps guide our approach to building the world's most trusted driver\cite{WaymoAdvancingVisionZero2025}" This evolutionary trajectory promises autonomous systems that harmonize human-like contextual understanding with robotic precision, fundamentally transforming personal mobility.

\subsection{LLM-Based Pedestrian Trajectory Prediction}
Recent studies leverage LLMs to transform pedestrian trajectory prediction from conventional geometric frameworks into semantic reasoning systems. By converting trajectory coordinates into symbolic representations , integrating scene context and social interactions via natural language, LLMs decode spatio-temporal dynamics to anticipate future paths. This paradigm shift enables socially aware reasoning, significantly enhancing trajectory accuracy and behavioral interpretability in crowded environments.

\subsubsection{Unique Challenges and LLM-Driven Advantages}
Pedestrian trajectory prediction faces fundamental challenges including the generalization limits of hand-crafted heuristic rules in open-world scenarios (e.g., unseen social dynamics) \cite{zhao2025trajevo}, data dependency of deep learning models requiring intensive trajectory annotations \cite{huang2024gpt}, interpretability gaps in safety critical applications demanding transparent reasoning chains \cite{kim2025guide}, and numerical-textual misalignment when processing continuous coordinates \cite{bae2024can}.
LLMs address these through evolutionary optimization of behavioral rules to mitigate generalization failures\cite{zhao2025trajevo}, synthetic trajectory generation using demographic priors to reduce annotation dependency\cite{ju2025trajllm}, explainable natural language rationalization to resolve interpretability gaps\cite{kim2025guide}, and discrete representation of continuous coordinates (e.g., translating $[(x_t,y_t)]$→``[(1.23,4.56)]") \cite{bae2024can} coupled with motion semantic extraction (e.g., inferring ``linear movement" from sequences) \cite{chib2024lg} to bridge numerical-linguistic disparities, collectively enabling robust open-world prediction with verifiable reasoning.

\subsubsection{Problem Formulation and Input Representation}
Pedestrian trajectory frameworks exhibit formalized structural innovations: Inputs integrate historical coordinates with scene context \cite{zhao2025trajevo,kim2025guide}, often augmented by LLM-derived motion semantics (e.g., ``curved motion", ``linear motion") \cite{chib2024lg}, future intention clusters modeled via Gaussian mixes \cite{chib2024lg}, and textual scene descriptions $\mathcal{P}{\mathcal{I}}$ generated using BLIP-2 \cite{bae2024can}. These are encoded through QA templates $\mathcal{T}{forecast}$ defining the prediction as language tasks \cite{bae2024can}. Outputs encompass multimodal trajectories with uncertainty-quantifying probabilities $\hat{p}_i$ \cite{chib2024lg}. Representationally, continuous coordinates are discretized into strings using hierarchical brackets \cite{bae2024can}, optimized via BPE tokenizers merging multi-digit sequences while preserving decimal integrity \cite{bae2024can}, with rank-k SVD distilling $>$95\% motion semantics from raw trajectories \cite{chib2024lg}.

\subsubsection{LLM-based Modeling Approaches}
\paragraph{Prompt Engineering and Task Formulation}
Prompt Engineering and Task Formulation leverage structured linguistic frameworks to optimize trajectory prediction\cite{pauk2025mapping}. Dynamic prompting employs iterative refinement of behavioral queries (e.g., ``Pedestrian 0's future path?") within end-to-end QA templates, where historical trajectories form contextual inputs $\mathcal{P}{\mathcal{C}}$ and predicted coordinates are output as textual sequences $\mathcal{P}{\mathcal{A}}$ \cite{bae2024can}. Task reformulation decomposes prediction into goal-oriented subtasks (``target position → trajectory generation") \cite{kim2025guide} or sequential encoding via encoder-decoder architectures to mitigate error accumulation \cite{bae2024can}.

\paragraph{Social Interaction Modeling}
Social Interaction Modeling combines implicit and explicit strategies\cite{wang2024socialformer}. 
Implicit modeling uses auxiliary QA tasks (group detection $\mathcal{T}{group}$ and collision risk assessment $\mathcal{T}{col}$) jointly trained to reinforce social relationship understanding\cite{bae2024can}. While explicit physical constraints (e.g., spatial impedance) are not directly enforced\cite{ju2025trajllm}, scene descriptions (e.g., descriptions of crowded areas implying reduced speed) implicitly guide socially compliant predictions\cite{yang2025trajectory}.

\begin{table*}[htbp]
\centering
\caption{Open-Source LLM Models for Trajectory Prediction (2023-2025)}
\vspace{-2mm}
\label{tab:os_models} 
\begin{tabularx}{\textwidth}{clllll}
\toprule
\textbf{Year} & \textbf{Paper Titles} & \textbf{Task Types} & \textbf{Vision Models} & \textbf{LLM Models} & \textbf{Outputs} \\
\midrule
2023 & GPT-Driver \cite{mao2023gpt} & Vehicle Traj & N/A & GPT-3.5 & Multiple Output Modalities \\
2024 & LMDrive \cite{shao2024lmdrive} & End-to-End (Traj) & ResNet-50/PointPillars & LLaVA-v1.5-7B & Multiple Output Modalities \\
2024 & DriveGPT4 \cite{xu2024drivegpt4} & End-to-End (Traj) & CLIP & LLaMA-2 & Multiple Output Modalities \\
2024 & DriveLM \cite{sima2024drivelm} & Vehicle Traj & BLIP-2 & Flan-T5-XL & Multiple Output Modalities \\
2024 & LLM-Trajectory-Prediction \cite{munir2024exploring} & Vehicle Traj & N/A & GPT/LLaMA/Zephyr/Mistral\textsuperscript{\ddag} & Low-Level Control; Trajectory \\
2024 &  LMTraj-SUP \cite{bae2024can} & Pedestrian Traj & BLIP-2 & GPT-3.5/GPT-4/Seq2Seq & Multiple Output Modalities \\
2024 & LLM-Augmented-MTR \cite{zheng2024large} & Traj (All agents\textsuperscript{\#}) & GPT4-V & GPT-4 & Multiple Output Modalities \\
2024 & VisionTrap \cite{moon2024visiontrap} & End-to-End (Traj) & BEVDepth/CNN & BERT & Trajectory \\
2025 & AutoVLA \cite{zhou2025autovla} & End-to-End (Traj) & Qwen2.5-VL-72B & Qwen2.5-3B & Multiple Output Modalities \\
2025 & OpenEMMA \cite{xing2025openemma} & End-to-End (Traj) & YOLO3D & Multi-LLM\textsuperscript{*} & Multiple Output Modalities \\
2025 & ReCogDrive \cite{li2025recogdrive} & End-to-End (Traj) & InternViT (300M) & Qwen2.5-7B & Multiple Output Modalities \\
2025 & OpenDriveVLA \cite{zhou2025opendrivevla} & End-to-End (Traj) & ResNet-101 & Qwen2.5-Instruct-(0.5B/3B/7B) & Trajectory \\
2025 & Trajectory-LLM \cite{yang2025trajectory} & Vehicle Traj & N/A & LLaMA-7B & Trajectory \\
2025 & LC-LLM \cite{peng2025lc} & End-to-End (Traj) & N/A & LLaMA-2-13B-chat & Multiple Output Modalities \\
2025 & DriveMoE \cite{yang2025drivemoe} & End-to-End (Traj) & Paligemma-3B & Paligemma-3B & Low-Level Control; Trajectory \\
2025 & SafeAuto \cite{zhang2025safeauto} & Vehicle Traj & LanguageBind & Vicuna-1.5-7B & Multiple Output Modalities \\
2025 & TRAJEVO \cite{zhao2025trajevo} & Pedestrian Traj & N/A & Gemini 2.0 Flash & Trajectory \\
2025 & GUIDE-CoT \cite{kim2025guide} & Pedestrian Traj & CLIP/U-Net & T5-small & Trajectory \\
2025 & TrajLLM \cite{ju2025trajllm} & Pedestrian Traj & N/A & LLaMA-3.1-8B-Instruct/GPT-4o-mini & Multiple Output Modalities \\
\bottomrule
\end{tabularx}

\vspace{0.1em}
\begin{flushleft}
\small
\textsuperscript{\ddag} “GPT/LLaMA/Zephyr/Mistral” includes GPT-2, LLaMA-7B, LLaMA-7B-Chat, Zephyr-7B, and Mistral-7B.\\
\textsuperscript{\#} “All agents” indicates that the model supports multiple traffic participants, including vehicles, pedestrians, and cyclists.\\
\textsuperscript{*} “Multi-LLM” includes combinations such as Mistral-7B, LLaMA-3.2-11B, Qwen2-7B, and GPT-4o.\\
\end{flushleft}
\vspace{-4mm}
\end{table*}

\paragraph{Multimodal Fusion and Constrained Generation}
Multimodal Fusion and Constrained Generation unify heterogeneous data via cross-modal concatenation of trajectory texts $\mathcal{P}{\mathcal{S}}$ and image captions $\mathcal{P}{\mathcal{I}}$ into single linguistic sequences \cite{bae2024can}. Generation constraints enforce feasibility through physical boundaries \cite{kim2025guide}, while temperature scaling ($\tau$=0.7) diversifies multimodal outputs and beam search ($d$=2) refines deterministic paths, balancing diversity with precision\cite{bae2024can}.

\subsubsection{Symbolic Reasoning and Motion Enhancement}
LLMs advance the prediction of the trajectory of pedestrians through the dual mechanisms of symbolic abstraction and the enhancement of motion representation\cite{chib2024lg}. Symbolically, LLMs convert raw trajectories into interpretable structures, including statistical feedback loops analyzing prediction utility distributions\cite{zhao2025trajevo}, verbalized goal positions (e.g., ``Pedestrian 0 → (57, 95)") for CoT reasoning \cite{kim2025guide}, and motion descriptors (e.g., ``linear movement") extracted via tailored prompts \cite{chib2024lg}. These enable explicit social reasoning through auxiliary QA tasks modeling group affiliations and collision risks \cite{bae2024can}. For motion enhancement, physical constraints enforce spatial feasibility using distance decay functions \cite{ju2025trajllm}, while dual historical/future motion cue fusion captures behavioral consistency \cite{chib2024lg}. Crucially, dedicated tokenizers preserve numerical continuity by separating integer/decimal components (e.g., ``6.25" → [``6", ``.", ``25"])—maintaining spatiotemporal coherence during linguistic transformation \cite{bae2024can}.

\subsubsection{Adaptive Rule Evolution and Trajectory Refinement}
LLMs enable adaptive prediction frameworks through rule evolution and trajectory refinement\cite{zhao2025trajevo}. For rule evolution, reflex mechanisms dynamically optimize behavioral heuristics (where short-term reflexes compare heuristic performance while long-term reflexes accumulate effective patterns), and user-guided refinement adjusts trajectory directions through goal probability modulation\cite{zhao2025trajevo,kim2025guide}. Data-driven motion patterns are automatically induced via Gaussian Mixture clustering of future trajectories \cite{chib2024lg}, supported by dual prediction paradigms: zero-shot inference through prompt engineering and supervised training via multi-task question answering \cite{bae2024can}. For trajectory refinement, rank-k Singular Value Decomposition (SVD) preserves $>$ 95\% of core motion patterns with $<$ 5\% information loss by approximating trajectory matrices, while probabilistic outputs quantify prediction uncertainty via nearest-neighbor soft probabilities \cite{chib2024lg}. Real-time optimization is achieved through beam search for deterministic path generation and temperature scaling ($\tau$=0.7 optimal) for multimodal diversity control, collectively balancing accuracy and computational efficiency in dynamic environments \cite{bae2024can}. Table\ref{tab:llm pedestrians} provides a comparative overview of these LLM-based paradigms, contrasting their core concepts, relative advantages, and inherent limitations.

\begin{table*}[!h]
    \centering
    \caption{Computational Efficiency Comparison of open-source LLM-Based trajectory prediction methods}
    \vspace{-2mm}
    \label{tab:efficiency}
    \begin{tabularx}{\linewidth}{lllp{2cm} p{2.5cm} l l}
        \toprule
        \textbf{Methods} & 
        \textbf{Datasets} &
        \textbf{Parameter Size} &
        \textbf{Inference Speed} & 
        \textbf{Inference Resources} & 
        \textbf{Training Resources} &
        \textbf{Training Time}  \\
        
        \midrule
        
        LMDrive\cite{shao2024lmdrive} & CARLA & 7B & - & - & 8*A100 & - \\
        VisionTrap\cite{moon2024visiontrap} & nuScenes-Text & - & 53ms & 1*3090Ti & - & - \\
        
        AutoVLA\cite{zhou2025autovla}& nuPlan Waymo & 3B & 500ms & - & 8*L40S  & - \\
        
        RecogDrive\cite{li2025recogdrive} & NAVSIM & 8B &- & - & 32*H20  & - \\
        OpenDriveVLA\cite{zhou2025opendrivevla} & nuScenes& 7B & - & - & 4*H100  &  2 days \\

        LC-LMM\cite{peng2025lc} & highD & - & 830ms & - & 8*A800 & 11 hours \\

        DriveMoE\cite{yang2025drivemoe} & CARLA & - & - & - & 4*H100  & 2 days \\
        
        DriveLM\cite{sima2024drivelm}& nuScenes & 3.955B & 6250ms & - &  8*V100     &  7 hours  \\
        
        LLM-Trajectory-Prediction\cite{munir2024exploring} & nuScenes & 7B & - & 1*3080Ti & 1*3080Ti & - \\
        
        SafeAuto\cite{zhang2025safeauto} & BDD-X DriveLM & 7B & - & - & 8*A6000  & - \\

        LMTraj-SUP\cite{bae2024can} & ETH-UCY SDD  & - & 18.3ms & - & 8*4090  & 4 hours \\

        TRAJEVO\cite{zhao2025trajevo} & ETH-UCY SDD & - & 0.65ms & - &  1*3090  & - \\

        GUIDE-CoT\cite{kim2025guide} & ETH-UCY & - & - & - & 1*3090Ti  & -\\
        \bottomrule
    \end{tabularx}
    \vspace{-4mm}
\end{table*}

\subsection{LLM-Based Traffic Participants Trajectory Prediction}
Current trajectory prediction research is moving from single-category modeling toward collaborative prediction for heterogeneous agents, where LLMs enable semantic reasoning between categories\cite{finet2025recent}. 
VisionTrap pioneers scene-centric multi-agent prediction\cite{moon2024visiontrap}: it employs vision-language models (BLIP-2) to generate agent descriptions refined by LLMs (GPT-4) with ground-truth maneuvers (e.g., ``stationary"), enforcing agent-text embedding alignment via debiased contrastive loss $\mathcal{L}{cl}$ ($\theta{th}=0.8$ filters false negatives); concurrently, it encodes multi-view images and HD maps into BEV features, extracting agent-specific cues (e.g., pedestrian gaze, turn signals) through deformable attention guided by predicted trajectories. This achieves a 43\% relative MR$_{10}$ reduction (0.56→0.32) and 27.56\% ADE$_{10}$ improvement (1.48→1.17) on nuScenes dataset, while processing 12 agents in 53ms. Complementarily, LLM-Augmented-MTR introduces cost-efficient semantic injection\cite{zheng2024large}: it visualizes vectorized maps and historical trajectories into Transportation Context Maps, prompting GPT4-V to output triple semantic descriptors—intentions (7 classes, e.g., ``straight-left"), affordances (5 classes, e.g., ``accelerate-allow"), and scenarios (4 classes, e.g., ``intersection")—as model inputs. By generalizing labels via nearest-neighbor algorithms with only 0.7\% LLM-augmented data, deployment costs plummet from 200k to 1.4k, yielding a 1.67\% multi-agent mAP gain on Waymo dataset.

\subsection{Open-Source LLM Models for Trajectory Prediction} 
Recent years have witnessed a rapid proliferation of open-source frameworks that integrate LLM into trajectory prediction. Table \ref{tab:os_models} catalogs 20 representative models with three key trends:
\begin{itemize}
    \item Lightweight Deployment: Emergence of sub-3B parameter models (e.g., Qwen2.5-VL-3B, T5-small) enabling real-time inference.
    \item Multimodal Fusion Dominance: 70\% of models (14/20) fuse vision-language-trajectory modalities, with CLIP/BLIP-2 as preferred encoders.
    \item Explanation-Centric Design: 65\% (13/20) generate \textit{ Trajectory + Natural Language} outputs for interpretability.
\end{itemize}
Beyond architectural trends, the computational efficiency and resource demands of these models are critical for their practical deployment. Table \ref{tab:efficiency}provides a detailed comparison of the inference speed and training resources required by a selection of  open-source methods. 

\begin{figure}[t!]
    \centering
    \includegraphics[scale=0.28]{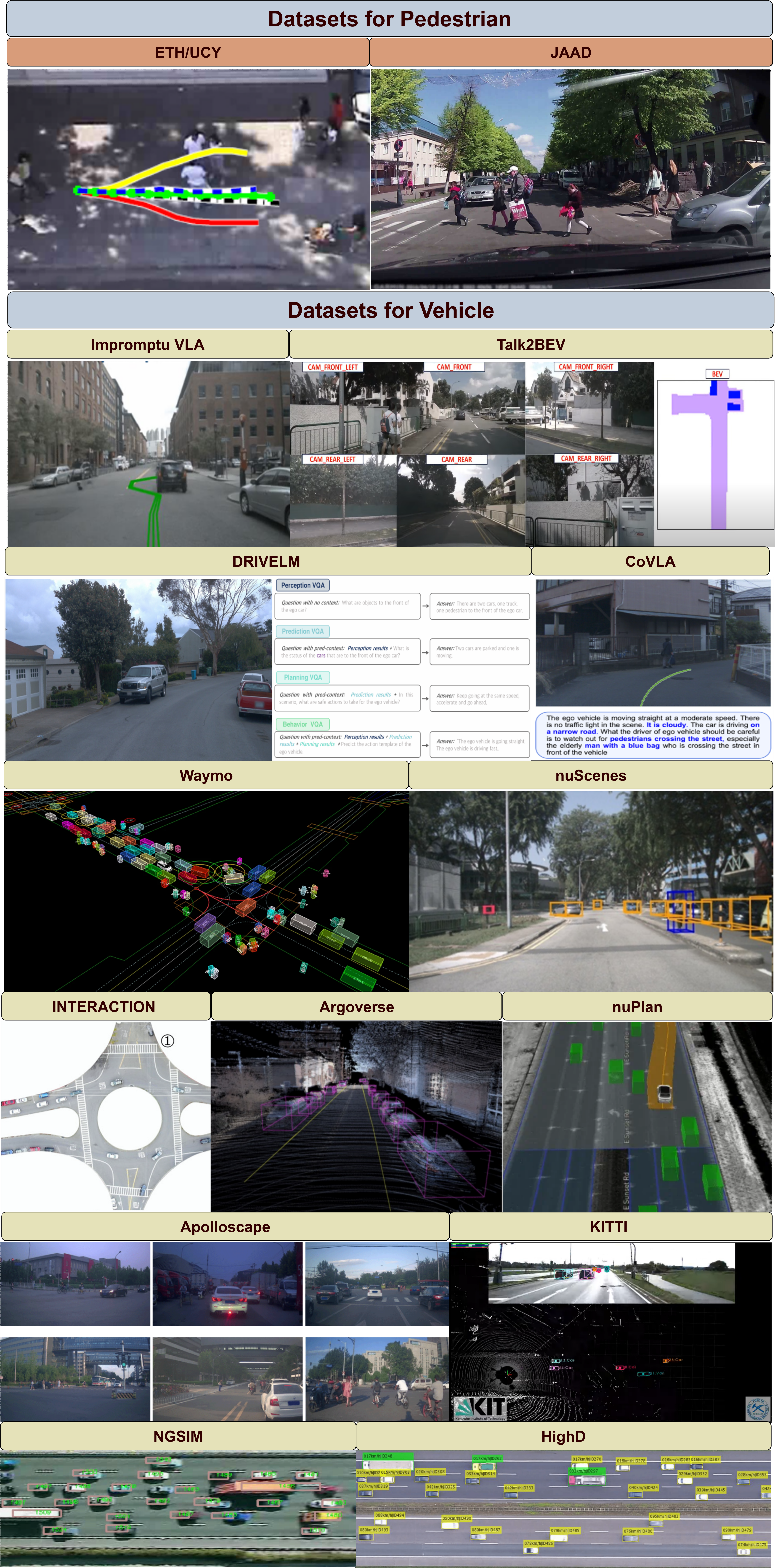}
    \vspace{-4mm}
    \caption{A Taxonomy of Datasets for Trajectory Prediction. Representative datasets for pedestrian prediction include ETH/UCY\cite{li2017scale} and JAAD\cite{kotseruba2016joint}, while vehicle prediction datasets include Impromptu VLA\cite{chi2025impromptu}, Talk2BEV\cite{choudhary2024talk2bev}, DRIVELM\cite{sima2024drivelm}, CoVLA\cite{arai2025covla}, Waymo\cite{sun2020scalability},nuScenes\cite{caesar2020nuscenes},INTERACTION\cite{zhan2019interaction},Argoverse\cite{chang2019argoverse},nuPlan\cite{caesar2021nuplan},Apolloscape\cite{huang2018apolloscape},KITTI\cite{geiger2013vision},NGSIM\cite{coifman2017critical} and HighD\cite{krajewski2018highd}.}
    \label{fig:dataset}
    \vspace{-4mm}
\end{figure}

\section{Experimental Benchmarks and Evaluation Metrics}

For the experimental evaluation of LLM-based trajectory prediction methods, selecting appropriate datasets is crucial. As summarized in Table \ref{tab:dataset-taxonomy}, existing datasets exhibit significant variation in their inherent suitability for LLM applications, primarily influenced by the availability and quality of language annotations and their specific scenario coverage. As visually categorized in Fig.~\ref{fig:dataset}, these benchmarks can be broadly grouped into those primarily targeting pedestrian prediction and those designed for vehicle trajectory prediction. This section benchmarks the performance of the reviewed LLM/MLLM approaches on prominent datasets representing various levels of LLM adaptation.

\begin{table*}[!h]
\centering
\caption{Taxonomy of Trajectory Prediction Datasets for LLM Applications}
\vspace{-2mm}
\label{tab:dataset-taxonomy}
\setlength{\tabcolsep}{4pt}
\fontsize{7.8}{8}\selectfont
\begin{tabularx}{\linewidth}{lXlllX}
\toprule
\textbf{Dataset}  & \textbf{Scenario Type} & \textbf{Language Annotation}  & \textbf{LLM Adaptation} & \textbf{Primary Applications}  \\
\midrule
NGSIM\cite{coifman2017critical}  & Highway  & \usym{2717} None  & Limited  & Physical Model Validation \\
highD\cite{krajewski2018highd}   & Highway  & \usym{2717} None  & Limited  & Vehicle Interaction Modeling \\
KITTI\cite{geiger2013vision}  & Urban Roads  & \usym{2717} None  & Limited  & Perception Benchmarking \\
ETH/UCY\cite{li2017scale}   & Pedestrian Zones  & \usym{2717} None  & Moderate  & Social Force Modeling \\
Apolloscape\cite{huang2018apolloscape}  & Urban Intersections    & \usym{2717} None  & Moderate  & E2E Trajectory Prediction \\
JAAD\cite{kotseruba2016joint}  & Pedestrian Crossings & \usym{2717} None  & High  & Pedestrian Intention Prediction \\
INTERACTION\cite{zhan2019interaction}  & Mixed Interaction Scenarios  & $\triangle$ Partial & High & Interactive Behavior Modeling \\
Argoverse\cite{chang2019argoverse}  & Urban Navigation  & \usym{2717} None & Moderate & Long-term Motion Forecasting \\
nuPlan\cite{caesar2021nuplan}  & Multi-city Urban  & $\triangle$ Map API & High & Motion Planning \\
Waymo\cite{sun2020scalability} & Multi-weather Urban  & $\triangle$ API Descriptions & High   & Industrial-grade SOTA Systems \\
nuScenes\cite{caesar2020nuscenes} & Multimodal Urban & \usym{2717} None  & Moderate & Multi-sensor Fusion \\
DRIVELM\cite{sima2024drivelm}  & Language-enhanced Scenes   & ✓ Manual  & \textbf{Optimal} & LLM-based Causal Reasoning \\
Talk2BEV\cite{choudhary2024talk2bev}  & BEV-Language Alignment   & ✓ Manual & \textbf{Optimal}  & Vision-Language Joint Training  \\
Impromptu  VLA\cite{chi2025impromptu} & Unstructured Roads  & $\triangle$ VLM-generated + Human-verified  & \textbf{Optimal}   & VLA Model Training \& Diagnostics  \\
CoVLA\cite{arai2025covla}  & Multi-condition Urban + Long-tail & ✓ Auto-generated +   Rule-constrained  & \textbf{Optimal}  & E2E VLA Planning \& Diagnostics\\
\bottomrule
\end{tabularx}
\end{table*}

\subsection{Vehicle Trajectory Prediction}
\paragraph{Datasets} The evaluation of vehicle trajectory prediction models predominantly utilizes large-scale, real-world driving datasets, which capture a wide variety of traffic scenarios, road geometries, and agent interactions. Key datasets include: (1) \textbf{nuScenes:} A comprehensive dataset for autonomous driving developed by Motional. It includes 1,000 driving scenes from Boston and Singapore, featuring a full sensor suite (6 cameras, 1 LiDAR, 5 RADAR). The dataset is richly annotated with 3D bounding boxes for 23 object classes and provides detailed map information, making it a standard benchmark for perception and prediction tasks\cite{caesar2020nuscenes}. (2) \textbf{Waymo Open Motion Dataset (WOMD):} Released by Waymo, this is one of the largest and most diverse datasets for motion forecasting. It contains over 100,000 scenes, each 20 seconds long, captured at 10Hz in various locations and weather conditions across the United States. The dataset provides detailed 3D labels for vehicles, pedestrians, and cyclists, along with high-resolution map data, making it ideal for studying complex interactions and long-term prediction\cite{sun2020scalability}. (3) \textbf{nuPlan:} Another dataset from Motional, nuPlan is specifically designed for motion planning. It contains over 1,500 hours of human driving data from four cities. Its large scale and focus on planning-centric metrics make it an invaluable resource for developing and validating closed-loop prediction and planning systems\cite{caesar2021nuplan}.
\begin{table*}[t]
\centering
\caption{Performance Comparison of Vehicle Trajectory Prediction Models on the nuScenes. Lower values indicate better performance.}
\vspace{-2mm}
\label{tab:exp-vehicle}
\scalebox{1}{
\begin{tabular}{lccccccccc}
\toprule
\multirow{2}{*}{\textbf{Method}} & \multirow{2}{*}{\textbf{Ego Status}} & \multicolumn{4}{c}{\textbf{L2 (m)}} & \multicolumn{4}{c}{\textbf{Collision (\%)}} \\
\cmidrule(lr){3-6} \cmidrule(lr){7-10}
& & \textbf{1s} & \textbf{2s} & \textbf{3s} & \textbf{Avg.} & \textbf{1s} & \textbf{2s} & \textbf{3s} & \textbf{Avg.} \\
\midrule
\multicolumn{10}{l}{\cellcolor[rgb]{0.957,0.957,0.957}\textit{Deep-Learning Method}} \\
UniAD \cite{huPlanningorientedAutonomousDriving2023} & & 0.2 & 0.42 & 0.75 & 0.46 & 0.02 & \textbf{0.25} & 0.84 & 0.37 \\
VAD-Base \cite{jiangVADVectorizedScene2023} & & 0.17 & 0.34 & 0.6 & 0.37 & 0.04 & 0.27 & \textbf{0.67} & \textbf{0.33} \\
AD-MLP \cite{zhaiRethinkingOpenLoopEvaluation2023} & & \textbf{0.15} & \textbf{0.32} & 0.59 & \textbf{0.35} & \textbf{0} & 0.27 & 0.85 & 0.37 \\
BEV-Planner \cite{liEgoStatusAll2024} & & 0.3 & 0.52 & 0.83 & 0.55 & 0.1 & 0.37 & 1.3 & 0.59 \\
BEV-Planner++ \cite{liEgoStatusAll2024} & & 0.16 & \textbf{0.32} & \textbf{0.57} & \textbf{0.35} & \textbf{0} & 0.29 & 0.73 & 0.34 \\
\hline
\multicolumn{10}{l}{\cellcolor[rgb]{0.957,0.957,0.957}\textit{LLM-based Method}} \\
DriveVLM \cite{tianDriveVLMConvergenceAutonomous2024} & & 0.18 & 0.34 & 0.68 & 0.4 & 0.1 & 0.22 & 0.45 & 0.27 \\
DriveVLM-Dual \cite{tianDriveVLMConvergenceAutonomous2024} & & 0.15 & 0.29 & 0.48 & 0.31 & 0.05 & \textbf{0.08} & \textbf{0.17} & \textbf{0.1} \\
EMMA \cite{hwangEMMAEndtoEndMultimodal2024} & & 0.14 & 0.29 & 0.54 & 0.32 & - & - & - & - \\
OmniDrive \cite{wangOmniDriveHolisticVisionLanguage2025} & & 0.14 & 0.29 & 0.55 & 0.33 & \textbf{0} & 0.13 & 0.78 & 0.3 \\
OpenEMMA \cite{xingOpenEMMAOpenSourceMultimodal2025} & & 1.45 & 3.21 & 3.76 & 2.81 & - & - & - & - \\
LightEMMA \cite{qiaoLightEMMALightweightEndtoEnd2025a} & & 0.28 & 0.93 & 2.02 & 1.07 & - & - & - & - \\
SOLVE-VLM \cite{chenSOLVESynergyLanguageVision2025a} & & \textbf{0.13} & \textbf{0.25} & \textbf{0.47} & \textbf{0.28} & \textbf{0} & 0.16 & 0.43 & 0.2 \\
\bottomrule
\end{tabular}
}
\vspace{-4mm}
\end{table*}

\paragraph{Metrics} The performance of vehicle trajectory prediction models is typically assessed using the following metrics:
\begin{enumerate}
    \item L2 Distance: This metric measures the Euclidean distance between the predicted trajectory's endpoint and the ground truth endpoint at a specific future time horizon (e.g., 1s, 2s, 3s). It provides a straightforward measure of prediction accuracy.
    \item Collision Rate: A critical safety metric, this measures the percentage of predicted ego-vehicle trajectories that result in a collision with other traffic agents or obstacles in the scene. It directly evaluates the model's ability to generate safe and feasible paths.
\end{enumerate}

\paragraph{Experimental Analysis} Table \ref{tab:exp-vehicle} offers a comparative analysis of deep learning-based and LLM-based methods for vehicle trajectory prediction, evaluated on L2 distance (m) and collision rate (\%) at 1, 2, and 3-second horizons.

The results clearly indicate a trend where LLM-based methods are advancing the state-of-the-art in vehicle trajectory prediction, particularly in enhancing safety.

\textbf{Superior Safety Performance:} The most significant insight is the superior performance of LLM-based methods in reducing collision rates. DriveVLM-Dual \cite{tianDriveVLMConvergenceAutonomous2024} and SOLVE-VLM \cite{chenSOLVESynergyLanguageVision2025a} stand out, achieving average collision rates of just 0.10\% and 0.20\%, respectively. This is a substantial improvement over the best-performing deep learning methods like VAD-Base (0.33\%) \cite{jiangVADVectorizedScene2023} and BEV-Planner++ (0.34\%) \cite{liEgoStatusAll2024}. This suggests that the semantic reasoning and world knowledge embedded in LLMs enable them to better anticipate and avoid potential conflicts, a critical capability for autonomous driving.

\textbf{Competitive Accuracy:} In terms of L2 distance, top-tier LLM-based models are highly competitive with and often surpass traditional deep learning approaches. SOLVE-VLM achieves the lowest average L2 error (0.28 m), outperforming all listed deep learning methods. Models like DriveVLM-Dual (0.31 m) and EMMA (0.32 m) \cite{hwangEMMAEndtoEndMultimodal2024} also demonstrate state-of-the-art accuracy. This indicates that integrating language-based reasoning does not come at the cost of geometric precision.

\textbf{Long-Horizon Improvement:} The advantages of LLM-based methods become more pronounced at longer prediction horizons. For instance, DriveVLM-Dual reduces the 3s collision rate to 0.17\%, a more than four-fold reduction compared to the 0.67-0.85\% rates of most deep learning methods. This highlights the strength of LLMs in complex, long-term reasoning, moving beyond simple pattern extrapolation.

\begin{table*}[t]
\centering
\caption{Performance Comparison (minADE/minFDE) of Pedestrian Trajectory Prediction Methods on the ETH-UCY Dataset.}
\label{tab:exp-pedestrian}
\scalebox{1}{
\begin{tabular}{lcccccc}
\toprule
\textbf{Method} & \textbf{ETH} & \textbf{Hotel} & \textbf{Univ} & \textbf{Zara1} & \textbf{Zara2} & \textbf{Avg} \\
\midrule
\multicolumn{7}{l}{\cellcolor[rgb]{0.957,0.957,0.957}\textit{Heuristical Method}} \\
ConstantAcc \cite{polychronopoulosSensorFusionPredicting2007} & 3.12/7.98 & 1.64/4.19 & 1.02/2.60 & 0.81/2.05 & 0.60/1.53 & 1.44/3.67 \\
CSCRCTR \cite{zhaiConstantSpeedChanging2014} & 2.27/4.61 & 1.03/2.18 & 1.35/3.12 & 0.96/2.12 & 0.90/2.10 & 1.30/2.83 \\
CVM \cite{schollerWhatConstantVelocity2020} & 1.01/2.24 & 0.32/0.61 & 0.54/1.21 & 0.42/0.95 & 0.33/0.75 & 0.52/1.15 \\
CVM-S \cite{schollerWhatConstantVelocity2020} & \textbf{0.92}/\textbf{2.01} & \textbf{0.27}/\textbf{0.51} & \textbf{0.53}/\textbf{1.17} & \textbf{0.37}/\textbf{0.77} & \textbf{0.28}/\textbf{0.63} & \textbf{0.47}/\textbf{1.02} \\
CTRV \cite{taoComparativeEvaluationKalman2021} & 1.62/3.64 & 0.72/1.09 & 0.71/1.59 & 0.65/1.50 & 0.48/1.10 & 0.84/1.78 \\
\hline
\multicolumn{7}{l}{\cellcolor[rgb]{0.957,0.957,0.957}\textit{Deep-Learning Method}} \\
Social-LSTM \cite{alahiSocialLSTMHuman2016} & 1.09/2.35 & 0.79/1.76 & 0.67/1.40 & 0.56/1.17 & 0.72/1.54 & 0.77/1.64 \\
Social-GAN \cite{guptaSocialGANSocially2018} & 0.87/1.62 & 0.67/1.37 & 0.76/1.52 & 0.35/0.68 & 0.42/0.84 & 0.61/1.21 \\
STGAT \cite{huangSTGATModelingSpatialTemporal2019} & 0.65/1.12 & 0.35/0.66 & 0.52/1.10 & 0.34/0.69 & 0.29/0.60 & 0.43/0.83 \\
Social-STGCNN \cite{mohamedSocialSTGCNNSocialSpatioTemporal2020} & 0.64/1.11 & 0.49/0.85 & 0.44/0.79 & 0.34/0.53 & 0.30/0.48 & 0.44/0.75 \\
PECNet \cite{mangalamItNotJourney2020} & 0.61/1.07 & 0.22/0.39 & 0.34/0.56 & 0.25/0.45 & 0.19/0.33 & 0.32/0.56 \\
AgentFormer \cite{yuanAgentFormerAgentAwareTransformers2021} & 0.46/0.80 & 0.14/0.22 & 0.25/0.45 & 0.18/0.30 & 0.14/0.24 & 0.23/0.40 \\
Trajectron++ \cite{salzmannTrajectronDynamicallyFeasibleTrajectory2021} & 0.61/1.03 & 0.20/0.28 & 0.30/0.55 & 0.24/0.41 & 0.18/0.32 & 0.31/0.52 \\
MemoNet \cite{xuRememberIntentionsRetrospectiveMemorybased2022} & 0.41/0.61 & \textbf{0.11}/\textbf{0.17} & 0.24/0.43 & 0.18/0.32 & 0.14/0.24 & 0.21/0.35 \\
MID \cite{guStochasticTrajectoryPrediction2022} & 0.57/0.93 & 0.21/0.33 & 0.29/0.55 & 0.28/0.50 & 0.20/0.37 & 0.31/0.54 \\
GP-Graph \cite{baeLearningPedestrianGroup2022} & 0.43/0.63 & 0.18/0.30 & 0.24/0.42 & 0.17/0.31 & 0.15/0.29 & 0.23/0.39 \\
NPSN \cite{baeNonProbabilitySamplingNetwork2022} & \textbf{0.36}/0.59 & 0.16/0.25 & 0.23/0.39 & 0.18/0.32 & 0.14/0.25 & 0.21/0.36 \\
SocialVAE \cite{xuSocialVAEHumanTrajectory2022} & 0.41/0.58 & 0.13/0.19 & \textbf{0.21}/\textbf{0.36} & 0.17/0.29 & 0.13/\textbf{0.22} & 0.21/0.33 \\
EigenTrajectory \cite{baeEigenTrajectoryLowRankDescriptors2023} & \textbf{0.36}/\textbf{0.53} & 0.12/0.19 & 0.24/0.43 & 0.19/0.33 & 0.14/0.24 & 0.21/0.34 \\
LED \cite{maoLeapfrogDiffusionModel2023} & 0.39/0.58 & \textbf{0.11}/\textbf{0.17} & 0.26/0.43 & 0.18/\textbf{0.26} & 0.13/\textbf{0.22} & 0.21/0.33 \\
MoFlow \cite{fuMoFlowOneStepFlow2025} & 0.40/0.57 & \textbf{0.11}/\textbf{0.17} & 0.23/0.39 & \textbf{0.15}/\textbf{0.26} & \textbf{0.12}/\textbf{0.22} & \textbf{0.20}/\textbf{0.32} \\
\hline
\multicolumn{7}{l}{\cellcolor[rgb]{0.957,0.957,0.957}\textit{LLM-based Method}} \\
LMTraj-SUP \cite{bae2024can} & 0.41/0.51 & \textbf{0.12}/0.16 & \textbf{0.22}/\textbf{0.34} & 0.20/0.32 & 0.17/0.27 & \textbf{0.22}/0.32 \\
TRAJEVO \cite{zhao2025trajevo} & 0.47/0.78 & 0.17/0.31 & 0.52/1.10 & 0.36/0.77 & 0.28/0.58 & 0.36/0.71 \\
GUIDE-CoT \cite{kim2025guide} & \textbf{0.38}/\textbf{0.43} & 0.13/\textbf{0.15} & 0.34/0.48 & \textbf{0.19}/\textbf{0.29} & \textbf{0.16}/\textbf{0.21} & 0.24/\textbf{0.31} \\
\bottomrule
\end{tabular}
}
\end{table*}

\subsection{Pedestrian Trajectory Prediction}
\paragraph{Datasets} For pedestrian trajectory prediction, research often focuses on datasets capturing complex, interactive social behaviors in crowded, unstructured environments. ETH-UCY is a collection of two widely-used datasets, ETH and UCY, which together contain five unique scenes (ETH, Hotel, Univ, Zara1, Zara2). These scenes feature pedestrians interacting in public spaces like university campuses and city squares. Despite being relatively small by modern standards, they remain a foundational benchmark for evaluating social interaction modeling due to the complex and nuanced behaviors they contain.

\paragraph{Metrics} The primary metrics for pedestrian trajectory prediction are minADE and minFDE, typically calculated over a set of K=20 predictions to evaluate the model's ability to capture the multimodality of human movement. The results are often reported as minADE / minFDE. minADE is the average L2 distance between the predicted trajectory and the ground truth, calculated over the best of the K predictions (the one with the lowest overall L2 distance). It is defined as:
\begin{equation}
    \text{minADE}_K = \min_{k \in \{1, \ldots, K\}} \frac{1}{T} \sum_{t=1}^{T} \left\| \hat{p}_k(t) - p_{gt}(t) \right\|_2
\end{equation}
where $T$ is the number of time steps, $\hat{p}_k(t)$ is the $k$-th predicted position at time $t$, and $p_{gt}(t)$ is the ground truth position.
Similar to minADE, minFDE measures the L2 distance between the predicted final position and the ground truth final position for the best of the K predicted trajectories. It is defined as:
\begin{equation}
    \text{minFDE}_K = \min_{k \in \{1, \ldots, K\}} \left\| \hat{p}_k(T) - p_{gt}(T) \right\|_2
\end{equation}

\paragraph{Experimental Analysis} As shown in Table \ref{tab:exp-pedestrian}, the evolution of pedestrian trajectory prediction methods, from simple heuristics to complex deep learning and LLM-based models, shows a clear and consistent improvement in accuracy.

\textbf{LLMs Pushing the Frontier}: LLM-based methods continue this trend, establishing a new state-of-the-art. LMTraj-SUP \cite{bae2024can} achieves an impressive average error of 0.22/0.32, which is competitive with the best deep learning models. Although TRAJEVO \cite{zhao2025trajevo} shows weaker performance, the strong results from LMTraj-SUP and GUIDE-CoT \cite{kim2025guide} indicate the high potential of this new paradigm. The ability of LLMs to interpret trajectory data within a rich semantic context, including the understanding of implicit social norms and intentions, enables more effective modeling of the stochastic and interactive nature of pedestrian behavior.

\textbf{Consistency Across Scenes}: Advanced deep learning and LLM-based methods generally show robust performance across all five scenes, which vary in density and interaction complexity. For example, LMTraj-SUP performs exceptionally well on the challenging ``Hotel" scene (0.12/0.16), matching or outperforming top deep learning models like MemoNet \cite{xuRememberIntentionsRetrospectiveMemorybased2022} and MoFlow \cite{fuMoFlowOneStepFlow2025}. This suggests that these newer models have better generalization capabilities.

\textbf{Semantic Reasoning is Key}: The success of models like GUIDE-CoT (0.24/0.31), which explicitly uses CoT reasoning, highlights that the improvements are not merely due to larger model capacity. Instead, the ability to perform step-by-step semantic reasoning about social situations is a key driver of performance. By formulating the prediction task as a language-based reasoning problem, LLM-based approaches can better handle the nuances of human behavior that are difficult to capture with purely geometric or statistical models.

\section{Discussion}
\subsection{Advantages of LLM-Based Trajectory Prediction}
\subsubsection{Semantic Reasoning for Enhanced Safety}
LLMs fundamentally transform trajectory prediction from pattern matching to context-aware reasoning. By integrating traffic rules (for example, ``yield to pedestrians at crosswalks"), social norms (e.g., pedestrian group behavior), and physical constraints through CoT prompting, LLMs generate trajectories that are not only statistically plausible but also compliant with rules and explainable\cite{liao2025cot}. This significantly reduces collision rates compared to traditional deep learning methods and addresses the black-box limitation of end-to-end neural networks \cite{chen2024end}.
\subsubsection{Generalization in Long-Tail Scenarios}
LLMs leverage pre-trained world knowledge to handle rare or unseen scenarios (e.g., irregular vehicles, construction zones)\cite{hegdeDistillingMultimodalLarge2025a}. For example, DriveVLM\cite{tian2024drivevlm} uses linguistic descriptions to recognize obscured objects, while retrieval-augmented methods such as RAG-Driver dynamically retrieve similar scenarios for zero-shot adaptation\cite{yuan2024rag}. This mitigates the data dependency of pure data-driven models\cite{li2022uqnet}.
\subsubsection{Multimodal Fusion Capability}
MLLMs unify heterogeneous inputs (LiDAR point clouds, camera images, HD maps) into a joint semantic space\cite{luo2025v2x}. Architectures like DiMA\cite{hegdeDistillingMultimodalLarge2025a} and DrivingGPT\cite{chen2024drivinggpt} align vectorized trajectories with visual language features via shared encoders, enabling holistic scene understanding crucial for complex interactions.

\subsection{Modular LLMs for Trajectory Prediction: Advantages over End-to-End Paradigms}
Despite the rise of end-to-end driving models, decoupling trajectory prediction with LLMs remains critical:
\subsubsection{Specialized Reasoning vs. Unified Optimization}
Trajectory prediction demands explicit spatiotemporal reasoning (e.g., multiagent interaction modeling\cite{seff2023motionlm}), while end-to-end systems prioritize control signal generation. LLMs allow focused optimization of semantic alignment (e.g., trajectory-language mapping via VQ-VAE\cite{chen2024drivinggpt}) without compromising planning latency. 
\subsubsection{Dynamic Knowledge Integration Efficiency}
The need for adaptable trajectory prediction systems motivates decoupled LLM-based architectures that enable efficient updates within constrained operational domains\cite{zhao2025trajevo}. Specifically, iMotion-LLM achieves targeted behavioral adaptation for predefined directional categories (e.g., stationary/straight/turn) through role-specific prompt templates that modify outputs without structural changes to its GameFormer backbone\cite{felemban2024imotion}. In addition, RAG-Driver implements training-free scenario adaptation via RA-ICL, dynamically sourcing expert demonstrations from analogous situations\cite{yuan2024rag}. This paradigm significantly reduces computational overhead compared to end-to-end model recalibration, as demonstrated by iMotion-LLM's parameter-efficient fine-tuning via LoRA updating only a minimal fraction of total parameters while maintaining core functionality\cite{felemban2024imotion}.

\subsubsection{Specialization for Spatiotemporal Reasoning}
Trajectory prediction necessitates explicit modeling of multi-agent interactions (e.g., joint distributions over interactive agent futures through autoregressive token decoding\cite{seff2023motionlm}) and kinematic scene constraints (e.g., efficient lane geometry compression via Bézier curve parameterization\cite{keysan2023can}). Decoupled language-model-based modules enable specialized optimization for these capabilities, as demonstrated by MotionLM’s latent-variable-free multimodal trajectory generation through discrete token sequences\cite{seff2023motionlm} while end-to-end systems prioritize control signal generation over structured scene representation.

\subsection{Challenges and Future Directions}
Although LLMs have shown considerable potential in trajectory prediction, a number of critical challenges remain that need to be addressed in future research.
\subsubsection{Persistent Challenges}
\paragraph{Computational Latency vs. Real-Time Requirements}
The significant latency of autoregressive decoding in LLMs often conflicts with typical vehicle control cycle requirements (e.g., 50 ms in DSDrive\cite{liu2025dsdrive}). While model compression techniques like distillation improve efficiency, they may lead to some loss in reasoning depth. Future work should focus on optimizing the trade-offs between latency and reasoning capabilities, such as dynamic computation allocation based on scenario complexity\cite{li2025recogdrive}.
\paragraph{Data-Scale and Distribution Shift}
LLMs demand large-scale paired data of trajectories and textual descriptions, yet real-world long-tail scenarios, such as heavy rain affecting LiDAR, remain a key source of prediction instability\cite{mcKinsey2024}. Although synthetic data generators such as GAIA-1\cite{wayve2023gaia1} can augment training, their physical realism is limited by simulation fidelity, introducing residual simulation-to-reality gaps (e.g., weather dynamics or sensor noise mismatches).
\paragraph{Robustness in Open-World Environments}
Adverse weather conditions and unstructured roadscapes pose substantial challenges to the robustness of autonomous driving systems. Sensors such as LiDAR and cameras degrade under rain, snow, fog or dust, leading to corrupted perceptual input\cite{dreissig2023survey}. Consequently, LLMs trained predominantly on clear-weather data may produce unsafe trajectories when exposed to such out-of-distribution conditions. Moreover, most models are designed under the assumption of structured urban environments, but real-world driving often involves construction zones, unpaved roads, parking lots, and scenarios with ambiguous or violated rules, where LLMs exhibit difficulty in interpreting intent or social norms\cite{chi2025impromptu}. 
\paragraph{Bias and Fairness}
Bias inherent in the training data of LLMs can manifest in their trajectory predictions, which raises significant concerns regarding fairness and safety\cite{gallegos2024bias}. For example, certain models may prioritize the safety of vehicle occupants over that of vulnerable road users, such as pedestrians. In addition, they may exhibit reduced prediction accuracy for less common agents, including those riding scooters or using wheelchairs. These limitations can introduce systematic biases that compromise the fairness and equity of the outcomes\cite{karim2025large}.

\subsubsection{Future Directions}
To address the aforementioned challenges, future research should evolve in the following directions:
\paragraph{Towards Ultra-Low Latency and Adaptive Inference}
Future research should focus on non-autoregressive decoding schemes for LLMs to generate trajectories in a single step, drastically reducing latency\cite{xu2025specee}. Furthermore, adaptive computation frameworks that allocate more computational resources to complex, high-risk scenarios (e.g., dense intersections) and simpler ones to straightforward scenes (e.g., highway cruising) will be crucial. The integration of specialized hardware with these algorithmic improvements could meet the stringent sub-50ms requirement.
\paragraph{Towards Foundation Models for Motion}
To mitigate data scarcity and the simulation-to-reality gap in LLM-based trajectory prediction, research should prioritize large-scale datasets encompassing long-tail scenarios, such as adverse weather, unstructured environments\cite{liao2025cot}. Leveraging LLMs’ strong sequence modeling and generalization capabilities, future work can focus on generating or augmenting trajectory data with high realism using generative diffusion models or GANs under LLM-guided synthesis. 
\paragraph{Towards World-Aware and Causally-Grounded Models}
Future work should advance LLMs from pattern matchers to reasoning-based world models by integrating robust multi-sensor representations resilient to noise, incorporating causal reasoning for agent intent and social interactions, embedding physical constraints for kinematic plausibility, and generating probabilistic multi-future predictions with explicit uncertainty quantification, enabling reliable open-world generalization.

\section{Conclusion}
This survey establishes that LLMs and MLLMs fundamentally transform trajectory prediction paradigms in autonomous driving. By translating continuous motion into discrete symbolic representations, LLMs enable a more explainable reasoning beyond traditional pattern matching. Key innovations include: (1) \textit{Trajectory-language mapping} through Bézier parameterization or VQ-VAE discretization, converting kinematics into semantically structured tokens; (2) \textit{Multimodal fusion architectures} (e.g., DiMA, DrivingGPT) that align vectorized motion states with visual-language features using shared scene encoders; and (3) \textit{Constraint-based reasoning} where natural language prompts explicitly encode traffic rules and physical constraints, reducing collision rates by 4 $\times$ versus baselines.  

For vehicle trajectory prediction, LLMs achieve state-of-the-art performance by decomposing interactions through CoT frameworks, progressing from intent recognition to risk-aware maneuver generation. Pedestrian prediction leverages LLMs' stochastic modeling capabilities, where behavior symbolization and social graph embeddings capture multimodal uncertainties. However, critical challenges persist in real-time deployment, including autoregressive decoding latency ($>$100ms) exceeding vehicle control cycles ($<$50ms), and distribution shifts in long-tail scenarios. Future work must prioritize latency-optimized distillation, causal invariance frameworks for safety certification, and physics-informed data augmentation to bridge simulation-to-reality gaps. The convergence of language-structured scene representation and spatio-temporal reasoning will ultimately enable human-like contextual awareness in trajectory prediction systems.

The list of abbreviations used in this survey is provided in Table\ref{tab:abbreviations}.

\begin{table}[htbp]
\scriptsize %
\centering
\caption{List of Abbreviations}
\label{tab:abbreviations}
\begin{tabular}{p{1.75cm}l}
\toprule
\textbf{Abbreviation} & \textbf{Full Term} \\
\midrule
LFMs & Large Foundation Models\\
LLMs & Large Language Models \\
MLLMs & Multimodal Large Language Models \\
HD & High Definition \\
RNNs & Recurrent Neural Networks \\
GNNs & Graph Neural Networks \\
CoT & Chain-of-Thought \\
IMM & Interacting Multiple Models \\
GP & Gaussian Processes \\
HMM & Hidden Markov Models \\
DBN & Dynamic Bayesian Networks \\
LSTMs & Long Short-Term Memory Networks \\
GRUs & Gated Recurrent Units \\
GANs & Generative Adversarial Networks \\
CVAEs & Conditional Variational Autoencoders \\
RL & Reinforcement Learning \\
DIRL & Deep Inverse Reinforcement Learning \\
GAIL & Generative Adversarial Imitation Learning \\
SFM & Social Force Model \\
VLMs & Vision-Language Models \\
RA-ICL & Retrieval-Augmented In-Context Learning \\
BEV & Bird's-Eye View \\
SFA & Semantic Fusion Attention \\
STG-Tokenizer & Spatial Temporal Graph Tokenizer \\
STG-Adapter & Spatial Temporal Graph Adapter \\
VQA & Visual Question Answering \\
STG & Spatial-Temporal Graph \\
minADE & Minimum Average Displacement Error \\
minFDE & Minimum Final Displacement Error \\
SVD & Singular Value Decomposition \\
WOMD & Waymo Open Motion Dataset \\
ICCV & IEEE/CVF International Conference on Computer Vision \\
CVPR & IEEE/CVF Conference on Computer Vision and Pattern Recognition \\
ECCV & European Conference on Computer Vision \\
NeuralPS & Conference on Neural Information Processing Systems \\
AAAI & AAAI Conference on Artificial Intelligence \\

IROS & IEEE/RSJ International Conference on Intelligent Robots and System \\
ICRA & International Conference on Robotics and Automation \\
HRI & ACM/IEEE international conference on human-robot interaction \\
RA-L & IEEE Robotics and Automation Letters \\
TIV & IEEE Transactions on Intelligent Vehicles \\
ICLR & International Conference on Learning Representation \\
TAI & IEEE Transactions on Artificial Intelligence \\
WACV & IEEE/CVF Winter Conference on Applications of Computer Vision \\
WWW & Companion Proceedings of the ACM on Web Conference \\
CTR & Communications in Transportation Research \\
TITS & IEEE Transactions on Intelligent Transportation Systems \\
\bottomrule
\end{tabular}
\end{table}

\bibliographystyle{IEEEtran}
\bibliography{references}

\end{document}